\documentclass[sigconf,10pt]{acmart}

\usepackage{caption}
\usepackage{graphicx}
\usepackage{tabularx} 
\usepackage[normalem]{ulem}
\usepackage{makecell} 
\usepackage{xspace}
\usepackage{booktabs}
\usepackage{multirow}
\usepackage{cleveref} 
\usepackage{lipsum}
\usepackage{enumitem}
\usepackage{amsmath}
\usepackage{float}
\usepackage{adjustbox}
\setlist[itemize]{noitemsep, topsep=0pt}
\usepackage{adjustbox}
\usepackage{subcaption}
\usepackage[normalem]{ulem}
\usepackage{algorithmicx}
\usepackage{algorithm}     
\usepackage{algpseudocode} 
\usepackage{amsmath}       
\usepackage{booktabs}   
\usepackage{pifont}     
\usepackage{xcolor}     
\usepackage{colortbl}   
\usepackage{graphicx}   

\usepackage{amssymb}

\newcommand{\cmark}{\ding{51}}%
\newcommand{\xmark}{\ding{55}}%
\definecolor{Gray}{gray}{0.9} 

\usepackage{soul}
\usepackage{hyperref}
\hypersetup{
    colorlinks=true,
    linkcolor=blue,
    filecolor=magenta,
    urlcolor=cyan,
}

\newcommand{\jl}[1]{{\color{purple}[JL: #1]}}

\newcommand{\system}{\textit{IROS}\xspace}

\begin{document}

\title[]{\system: A Dual-Process Architecture for Real-Time VLM-Based Indoor Navigation}


\author{Joonhee Lee}
\email{neo81389@yonsei.ac.kr}
\affiliation{%
 \institution{Yonsei University}
 \country{Seoul, South Korea}
}

\author{Hyunseung Shin}
\email{jay040730@yonsei.ac.kr}
\affiliation{%
 \institution{Yonsei University}
 \country{Seoul, South Korea}
}

\author{JeongGil Ko}
\email{jeonggil.ko@yonsei.ac.kr}
\affiliation{%
 \institution{Yonsei University}
 \country{Seoul, South Korea}
}

\begin{abstract}

Indoor mobile robot navigation requires fast responsiveness and robust semantic understanding, yet existing methods struggle to provide both. Classical geometric approaches such as SLAM offer reliable localization but depend on detailed maps and cannot interpret human-targeted cues (e.g., signs, room numbers) essential for indoor reasoning. Vision-Language-Action (VLA) models introduce semantic grounding but remain strictly reactive, basing decisions only on visible frames and failing to anticipate unseen intersections or reason about distant textual cues. Vision-Language Models (VLMs) provide richer contextual inference but suffer from high computational latency, making them unsuitable for real-time operation on embedded platforms. In this work, we present \system{}, a real-time navigation framework that combines VLM-level contextual reasoning with the efficiency of lightweight perceptual modules on low-cost, on-device hardware. Inspired by Dual Process Theory, \system{} separates fast reflexive decisions (System One) from slow deliberative reasoning (System Two), invoking the VLM only when necessary. Furthermore, by augmenting compact VLMs with spatial and textual cues, \system{} delivers robust, human-like navigation with minimal latency. Across five real-world buildings, \system{} improves decision accuracy and reduces latency by 66\% compared to continuous VLM-based navigation.

\end{abstract}

\settopmatter{printacmref=false}
\setcopyright{none}
\renewcommand\footnotetextcopyrightpermission[1]{}
\pagestyle{plain}

\setcopyright{none}
\makeatletter
\renewcommand\@formatdoi[1]{\ignorespaces}
\makeatother

\maketitle

\section{Introduction}
\label{sec:intro}

Indoor navigation is a fundamental capability for intelligent robotic systems deployed in human environments~\cite{raj24intelli, omer24semantic, zhao18p3loc}. From delivery robots in office buildings to service robots in hospitals and residential complexes, robots must interpret complex visual scenes, react to dynamic changes, and make safe decisions in real-time~\cite{baek2025ai}. Yet, reliable indoor navigation remains difficult as indoor spaces are visually diverse, cluttered with human-targeted cues (e.g., room numbers, signs, bulletin boards), and often structured in ways that demand semantic rather than purely geometric reasoning. A robot traversing hallways, corners, and intersections must process sensor data and act quickly, as any delay risks collisions, user discomfort, or task failure.

Traditional approaches rely on geometric pipelines such as SLAM~\cite{cadena2016past}; yet, these approaches face distinct hardware and environmental constraints. LiDAR-based SLAM systems~\cite{bresson2017simultaneous} are often restricted by high sensor costs, preventing affordable wide-scale use. RGB-based alternatives, while accessible, lack deep knowledge of environment characteristics.
Beyond sensor-centric limitations, geometric navigation fundamentally lacks the capacity to interpret visual semantic cues, providing only metric representations of the environment.

\begin{figure}[!t]
\centering
\includegraphics[width=0.99\linewidth]{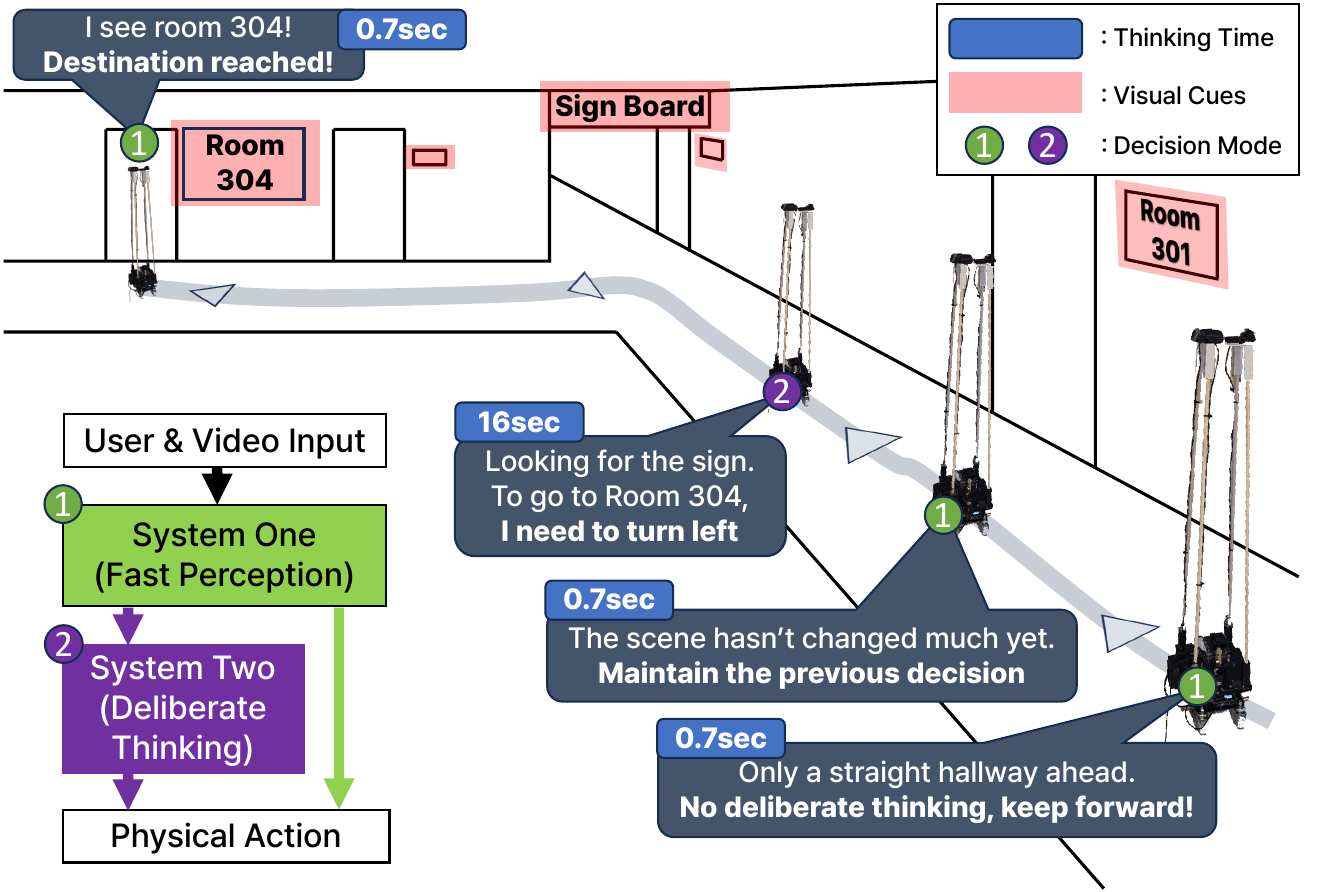}
\vspace{-2ex}
\caption{Conceptual example of \system{} with semantic-level navigational input, ``go to room 304.''}
\vspace{-2ex}
\label{fig:conceptual_example}
\end{figure}

Vision-Language-Action (VLA) models have recently emerged as an alternative approach that couples visual perception with language-conditioned control. While VLAs~\cite{glossop2025cast, rt12022arxiv, rt22023arxiv, driess23palme} offer stronger semantic grounding compared to SLAM-based navigational pipelines, they remain fundamentally \textit{reactive} as their decisions are bound to objects and affordances visible in the current frame or short temporal window. Thus, VLAs struggle with complex navigation tasks, such as inferring locations of unseen targets based on semantic cues, deducing a building's layout logic, or navigating towards a goal that requires long-horizon spatial reasoning.

More general Large Language Models (LLMs) and Vision-Language Models (VLMs), in contrast, possess broad world knowledge and can reason contextually beyond what is immediately visible, enabling inference about goals, semantics, and partially observed spaces. Their intuitive, context-aware reasoning makes them appealing as the cognitive core of next-generation robots. However, deploying such models introduces a critical bottleneck of \textit{computational latency}~\cite{kong24arise, bin24coacto, rastikerdar24cactus, jia22codl, zhang21elf, wang22melon, xu24edgellm, gim2022sage}. VLM inference is slow and computationally heavy, especially on embedded hardware, and their computation delays can break real-time responsiveness~\cite{xu24edgellm, li2024survey}. A robot that repeatedly ``pauses to think'' for long periods becomes impractical and unsafe.

These limitations reveal an important research opportunity. While VLMs offer the semantic and contextual reasoning needed for complex indoor navigation, they cannot satisfy the strict real-time constraints required for safe physical operation. This motivates a design paradigm that couples the cognitive strengths of VLMs with the speed and reactivity of lightweight perceptual modules, enabling robots to think like humans \textit{without} sacrificing real-time performance.

We propose \system{}, a real-time indoor navigation framework that preserves the intuitive, context-aware reasoning capabilities of VLMs while meeting the latency demands of physical deployments. As Figure~\ref{fig:conceptual_example} illustrates, \system{} adopts a dual-pathway architecture inspired by Dual Process Theory~\cite{kahneman2011thinking, Ganapini25llavar, han24dualprocessvlaefficient}, consisting of a fast, reflexive pathway for frequent, low-ambiguity decisions (System One) and a slow, deliberative pathway for complex or uncertain cases (System Two). Designed to operate fully on low-cost, on-device hardware without external connectivity, \system{} further improves efficiency through Spatial \& Textual Information Augmentation by supplying the VLM with structured environmental cues, enhancing its decision-making accuracy.


Evaluations in five real environments show that \system{} reduces average travel time by 66\% while improving decision fidelity over VLM-only baselines. Furthermore, System One handles more than half of maneuvering decisions with sub-second latency, and our Spatial/Textual Information Augmentation improves the VLM's maneuver decision accuracy from 48.2\% to 64.3\%. These results demonstrate that \system{} enables robust, real-time navigation using compact VLMs.

The key contributions of this work are as follows:
\begin{itemize}[leftmargin=*]
    \item \textbf{Dual-Process Navigation Architecture.}
    We introduce a decoupled architecture for VLM-based robotic navigation comprising System One (fast perception) and System Two (deliberative reasoning). System One handles immediate, reflex-like decisions using lightweight vision modules, while System Two invokes a VLM only when higher-level reasoning is needed. This separation preserves VLM semantic intelligence while ensuring real-time response.

    \item \textbf{Conditional Inference for Efficient Computation.}
    We design \textit{Key Frame Compare} and \textit{Vision–Condition Matching} modules that trigger VLM reasoning only upon meaningful environmental change. This conditional execution reduces latency and improves system responsiveness.

    \item \textbf{Spatial and Textual Information Augmentation.}
    Compact VLMs often lack spatial awareness. \system{} augments VLM prompts with explicit spatial descriptors derived from geometric segmentation, and incorporates an OCR pipeline to capture distant signboards and textual cues that are easily missed by standard vision encoders.

    \item \textbf{End-to-End Implementation and Real-World Evaluation.}
    We implement \system{} on a mobile robot and extensively evaluate it across diverse real-world indoor environments. Our results demonstrate that \system{} achieves human-like decision making with low latency, effectively bridging the gap between the reasoning capability of VLMs and the demands of practical robotic navigation.
\end{itemize}

\section{Background and Motivation}
\label{sec:background_motivation}

\subsection{RL and VLMs as Decision Makers}

A core challenge for mobile robots operating as embodied agents is completing the \textit{perception–decision–action} loop reliably in real-world environments that are noisy, dynamic, and structurally unbounded. Designing a decision-making module that is both robust and efficient under such conditions remains an open problem.

Two dominant paradigms have emerged for this decision-making role. The first is Reinforcement Learning (RL), which learns policies through extensive trial-and-error interactions. While RL achieves impressive performance in well-scoped tasks, its policies often generalize poorly. Robots trained in simulation or limited real-world datasets frequently fail when encountering unfamiliar scenarios and remain highly vulnerable to the well-documented ``sim-to-real'' gap~\cite{akkaya2019solving}.

The second paradigm employs large, pre-trained Vision-Language Models (VLMs) as reasoning engines. Recent work, such as VLFM~\cite{yokoyama2024vlfm}, NavVLM~\cite{yin2025navvlm} and OmniNav~\cite{xue25omni}, and others \cite{zhao24overnav, wang24goat, liu24ver, zhou24navid, shah22lmnav}, demonstrates that VLMs can interpret unstructured scenes, integrate visual and textual cues, and select contextually appropriate actions using their extensive world knowledge. Yet this strength comes with a major limitation: VLM inference is computationally expensive and introduces substantial latency. Moreover, sustaining multi-step reasoning often necessitates recursive or hierarchical prompting, further increasing inference time.

In short, RL offers fast execution but poor adaptability, whereas VLMs provide strong generalization and reasoning but lack real-time responsiveness. This fundamental trade-off motivates our development of a hybrid, dual-system architecture that integrates the complementary strengths of both paradigms to support real-time, context-aware decision making in autonomous robotic systems.

\subsection{Other Vision Models as Decision Makers}

VLMs generate textual reasoning from joint visual and textual inputs. Trained on massive multimodal datasets, they acquire an intuition-like ability to connect perceptual cues with plausible next actions, a reasoning capability far beyond what conventional vision-only models can provide.

\begin{table}[t]
\centering
\includegraphics[width=\columnwidth]{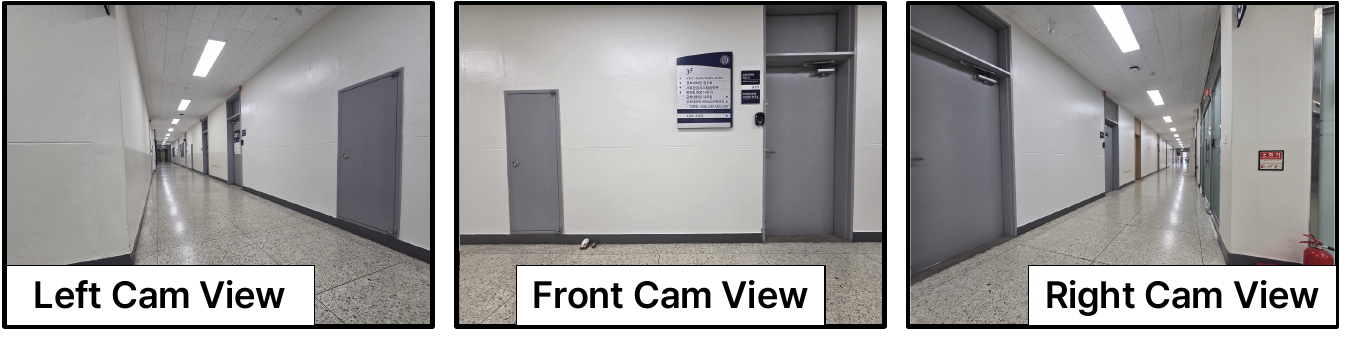}
\vspace{-1ex}

\resizebox{\columnwidth}{!}{%
\begin{tabular}{l c p{3.5cm} p{3.8cm}}
\toprule
\textbf{Feature / Method} & 
\textbf{\begin{tabular}[c]{@{}c@{}}YOLO-World\\ \cite{cheng2024yoloworld}\end{tabular}} & 
\multicolumn{1}{c}{\textbf{\begin{tabular}[c]{@{}c@{}}BLIP-2\\ \cite{li2023blip}\end{tabular}}} & 
\multicolumn{1}{c}{\cellcolor{Gray}\textbf{\begin{tabular}[c]{@{}c@{}}Gemma3 4B\\ \cite{team2024gemma}\end{tabular}}} \\ \midrule

Perception Capability & 
Object Detection & 
\multicolumn{1}{c}{Image Captioning} &
\multicolumn{1}{c}{\cellcolor{Gray}\textbf{Visual Reasoning}} \\ 

Visual Information Retrieving & 
$\triangle$ &
\multicolumn{1}{c}{\cmark} &
\multicolumn{1}{c}{\cellcolor{Gray}\textbf{\cmark}} \\ 

Reasoning \& Action Decision & 
\xmark & 
\multicolumn{1}{c}{\xmark} & 
\multicolumn{1}{c}{\cellcolor{Gray}\textbf{\cmark}} \\ \midrule

\textbf{Example Output} & 
\textit{N/A} & 
``Bulletin board ahead, hallway on the left and right'' & 
\cellcolor{Gray}\textbf{``Bulletin board ahead... A301-310 with the left arrow. To go to A303, turn left.''} \\ \bottomrule
\end{tabular}%
}
\caption{Comparison of perception and reasoning capabilities for different models with sample camera view.}
\vspace{-2ex}
\label{tab:vision_based_model_comparison}
\end{table}

To compare a VLM with vision-only alternatives, we perform functional comparisons with BLIP-2~\cite{li2023blip} and YOLO-World~\cite{cheng2024yoloworld}, which are the perception components used in previous work such as VLFM~\cite{yokoyama2024vlfm} and VL-Nav~\cite{du2025vlnav}. We examine these models on a hallway-intersection scenario requiring spatial reasoning. As Table~\ref{tab:vision_based_model_comparison} shows, YOLO-World is limited to detecting objects matching user queries and thus cannot answer open-ended navigational questions. On the other hand, BLIP-2 can describe the scene but lacks the reasoning necessary to convert that description into an action. In contrast, the Gemma3 4B VLM successfully interpreted the scene and selected the correct navigational direction.

These results highlight a fundamental limitation of traditional vision models: they can detect or describe, but cannot translate perception into correct action without an external reasoning assistance. VLM-level reasoning is thus essential for autonomous navigation in human-centric environments.

\subsection{On-Device Deployment Challenges}

For robots operating in the physical world, deploying the full intelligence stack on-device (e.g., NVIDIA Jetson) is essential to avoid network instability and privacy risks associated with cloud-based inference. However, on-device execution of large models introduces inference latency as a major bottleneck~\cite{zhu2024on_device_train}. In mobile robotics, latency manifests as extended ``thinking time,'' delaying actuation and increasing collision risk in dynamic environments.

Conventional model-optimization techniques, such as quantization, pruning, or edge–cloud offloading~\cite{kong24arise, bin24coacto, rastikerdar24cactus, jia22codl, zhang21elf, wang22melon}. primarily reduce model size rather than addressing the efficiency of the reasoning process itself. We argue that simply shrinking the model is insufficient; achieving true real-time performance requires a \textit{system-level architectural} redesign of the decision-making pipeline.

\subsection{Theoretical Foundation: \\Dual Process Theory}

To balance reasoning capability with computational efficiency, we draw inspiration from human cognition, specifically the \textbf{Dual Process Theory}~\cite{kahneman2011thinking}. This framework distinguishes between a fast and intuitive \textbf{System One} that makes rapid, simple, and computationally lightweight decisions, and a slow but deliberative \textbf{System Two} for making logical accurate and resource-intensive decisions.


This cognitive duality has motivated hierarchical robotic architectures such as CogDDN~\cite{huang25cogdnn}. Unlike prior approaches that primarily focus on layering sensor abstractions, we apply this dual-process principle directly to the navigation decision pipeline, explicitly separating high-level semantic reasoning from low-level spatial reflexes.

\subsection{Empirical Motivation: Addressing VLM Reasoning Redundancy}

Our preliminary experiments reveal a key inefficiency when using VLMs as the sole decision maker. In continuous navigation tasks such as traversing a straight hallway, the VLM (System Two) repeatedly generates the same decision across hundreds of consecutive frames.

\begin{figure}[!t]
\centering
\includegraphics[width=0.95\linewidth]{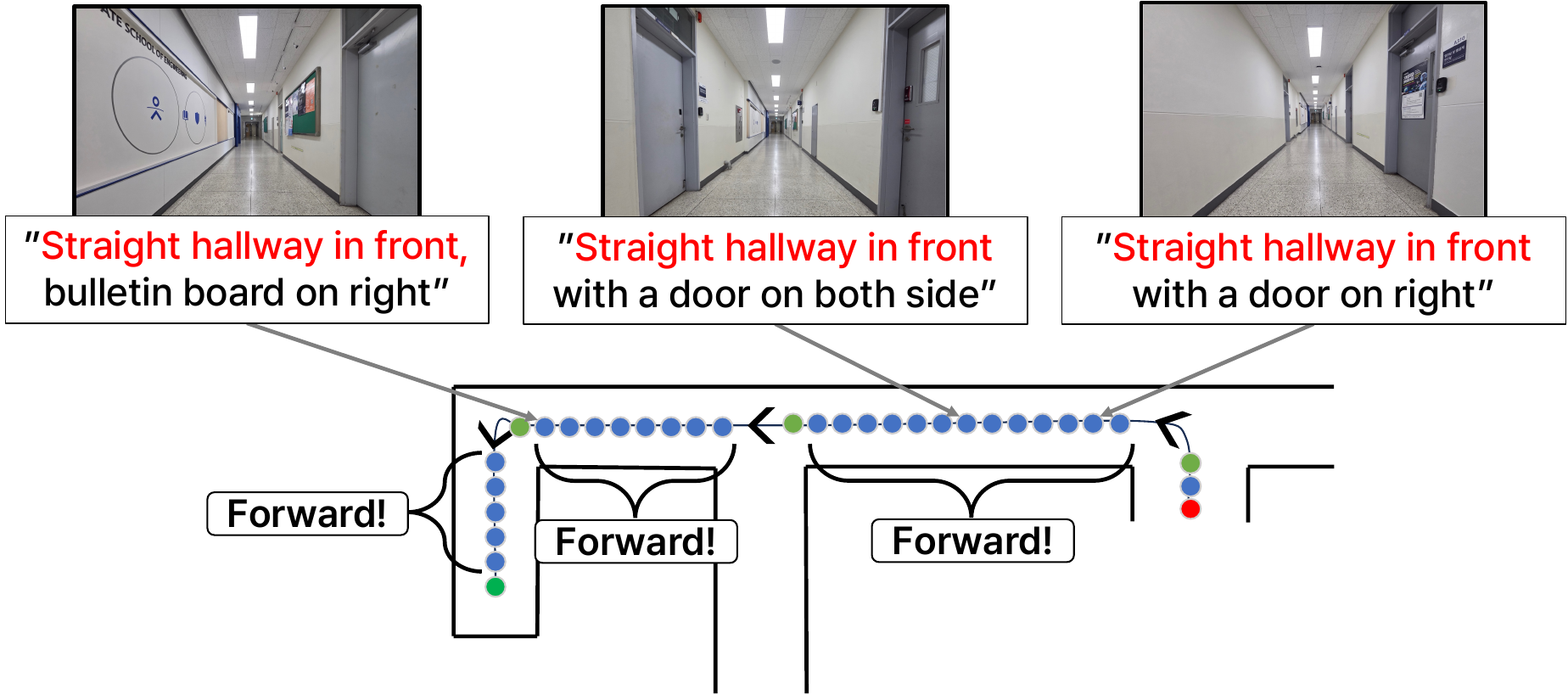}
\vspace{-2ex}
\caption{Example scenario of making unnecessary repetitive VLM calls on hallways.}
\label{fig:inefficiency_vlm_inference}
\end{figure}

As shown in Figure~\ref{fig:inefficiency_vlm_inference}, meaningful changes in decision outputs occur only near corners, intersections, or other structurally significant transitions. Straight-path regions, by contrast, produce near-identical reasoning outcomes, making continuous VLM invocation unnecessarily expensive. This redundancy suggests an opportunity for architectural optimization: if these repetitive, low-complexity scenarios are delegated to a lightweight System One mechanism, VLM inference can be reserved for moments that truly require semantic or contextual reasoning. This insight provides the empirical foundation for our proposed dual-system design.

\section{\system Design Overview}
\label{sec:System}

\begin{figure*}[!t]
\centering
\includegraphics[width=0.85\linewidth]{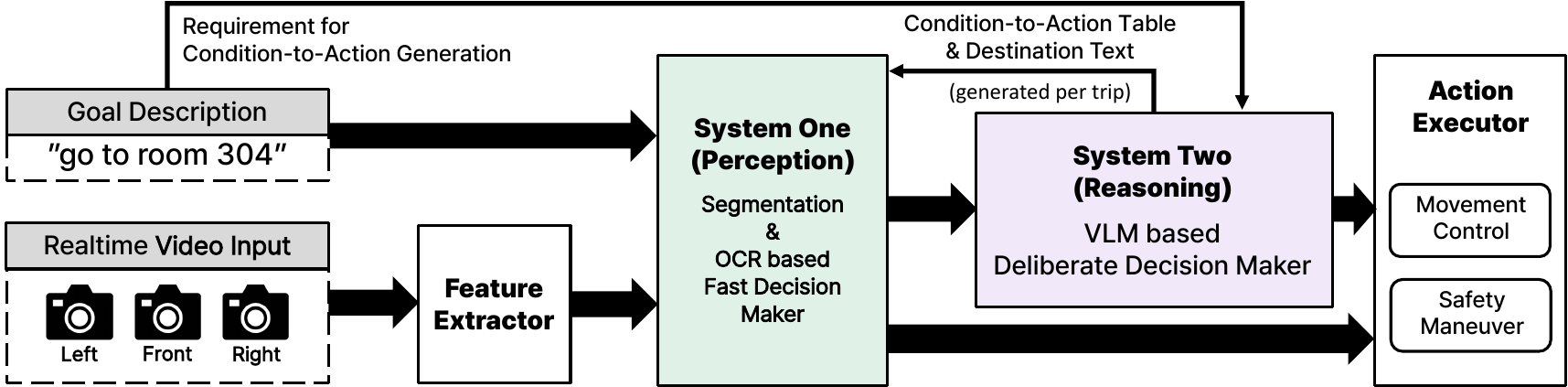}
\vspace{-2ex}
\caption{Overview of \system workflow.}
\label{fig:system_overview}
\end{figure*}

Building on the limitations of VLM-based navigation, a system-level redesign is needed to support \textit{real-time, latency-sensitive} robotic tasks. Our analysis shows that in continuous navigation (e.g., path following), VLMs frequently produce identical decisions across consecutive frames, resulting in computational redundancy and underutilized reasoning capability. Motivated by the \textit{Dual Process Theory}, we propose \system{}, a dual-system architecture that maximizes both real-time performance and GPU efficiency in vision–language decision making. Figure~\ref{fig:system_overview} provides a high-level overview.
Specifically, to achieve fast yet accurate decisions, \system{} separates \textbf{perception} (\textit{System One}) from \textbf{reasoning} (\textit{System Two}):

\vspace{2ex}

\noindent\textbf{System One (Perception)} acts as the ``fast decision'' pathway, focusing on spatial awareness and immediate response. To retain sufficient intelligence while ensuring rapid execution, it utilizes a precomputed \textit{Condition-to-Action} table to facilitate rapid \textit{vision–condition matching}, enabling real-time decisions without invoking the VLM. This mechanism efficiently handles common, low-ambiguity scenarios. Notably, the \textit{Condition-to-Action} table is generated by the VLM during initialization; the VLM analyzes the destination and surrounding environment to anticipate potential scenarios and pre-maps corresponding actions to each condition.

\vspace{2ex}

\noindent\textbf{System Two (Reasoning)} exploits a VLM at its core and serves as the ``slow but deliberate'' pathway, activated when System One cannot determine a unique action or when deeper semantic understanding is required. It performs (i) offline \textit{Condition-to-Action Table Generation} and (ii) real-time \textit{Additional Thinking}. The latter leverages VLMs to interpret complex visual cues (e.g., signs, room numbers, or bulletin boards), enabling context-aware decisions. Since compact on-device models lack spatial \& textual grounding, System One’s spatial \& textual analysis is injected into System Two’s prompt to improve reasoning accuracy.

This dual-system architecture enables \system{} to execute most decisions in real time while selectively invoking the VLM for high-level reasoning. By exposing configurable parameters, such as token budgets, similarity thresholds, and segmentation granularity, \system{} remains both robust across diverse indoor environments and flexible for deployment on various hardware platforms.

\section{Detailed Design of \system}
\label{sec:System}

Before diving into discussing the detailed design of \system{}, we start by introducing key terminologies used throughout this section to ensure reading clarity:

\begin{itemize}[leftmargin=*]
    \item \textbf{Condition:} A predefined textual descriptor of an environmental state (e.g., ``A straight hallway''), formulated during the \textit{Condition-to-Action Generation} phase. Each condition is associated with specific robot actions and stored in the \textit{Condition-to-Action} table.
    \item \textbf{Action:} A discrete low-level command that can be directly executed by the robot’s motion controller (e.g., \textit{Forward}, \textit{CW($N^\circ$)}, \textit{CCW($N^\circ$)}, \textit{Turn Back}).
\end{itemize}

\vspace{0.1cm}
These definitions form the foundational vocabulary for understanding how \system{} coordinates perception, reasoning, and actuation in the following subsections.

\subsection{System One: Fast Perception/Matching}

\begin{figure}[!t]
\centering
\includegraphics[width=0.95\linewidth]{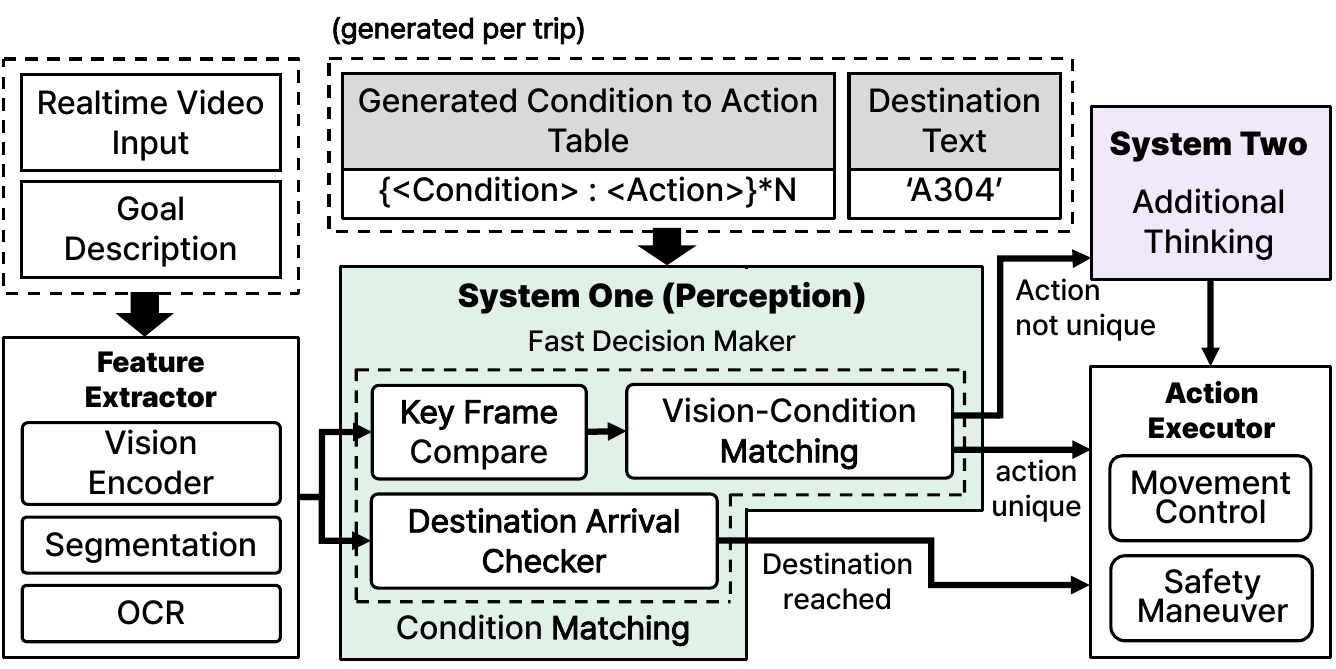}
\vspace{-2ex}
\caption{Detailed overview of System One}
\vspace{-2ex}
\label{fig:system_one_overview}
\end{figure}

\textbf{System One} functions as the rapid, reflexive component of the \system{} architecture, designed for real-time environmental interaction. To support these operations, \system{} relies on the \textit{Feature Extractor}, a composite module containing the Vision Encoder, Segmentation, and OCR components. System One selectively accesses the outputs from these sub-modules as needed for each processing stage. The overall workflow is illustrated in Figure~\ref{fig:system_one_overview} and proceeds as follows.

\begin{enumerate}[leftmargin=*]
    \item \textbf{Key Frame Compare (KFC):} Using embeddings from the Vision Encoder, this gating mechanism measures structural differences between the current frame and the last frame used for robot movement decision-making. It serves as a trigger, activating subsequent phases only when significant visual changes indicate a new state.
    
    \item \textbf{Condition Matching:} When activated, Condition Matching performs two parallel assessments. First, \textit{Destination Arrival Checking} computes vision-text similarity between the current view and Goal Description while verifying specific Destination Text. Second, \textit{Vision Condition Matching} generates spatial and textual scene descriptions and compares them against the \textit{Condition-to-Action} table via cosine similarity to identify a set of potential actions.
    
    \item \textbf{Action Execution or Escalation:} If a unique action is identified by the decision making process (or if the Destination Arrival Checking mode asserts an \texttt{Idle} state), the \textit{Action Executor} takes the appropriate action. If zero or multiple actions are found during the augmented condition matching phase, the situation is deemed ambiguous and escalated to System Two for \textit{Additional Thinking}.
\end{enumerate}

\subsubsection{Key Frame Compare}

To reduce unnecessary computation, \system{} incorporates \textit{Key Frame Compare (KFC)}, a lightweight gating mechanism that triggers navigational reasoning from System Two only when the visual scene meaningfully changes. KFC continuously monitors incoming frames from the Vision Encoder (VE; SigLIP~\cite{zhai2023sigmoid}; residing in the Feature Extractor) and compares them against the last inference-triggering frame. When structural differences exceed a threshold, System Two is activated, else, System One determines the navigation actions.

Instead of relying on a single-pooled embedding, which predominantly captures global semantics, \system{}'s KFC utilizes patch-level embeddings that represent the image as a grid of spatially corresponding local features. This design emphasizes spatial layout and structural details. For instance, while a `straight hallway' and a `hallway approaching a corner' may share the same global semantics, they exhibit distinct feature distributions in specific local patches (e.g., image edges that show hallway splits). By comparing patch-wise features, \system{} detects navigationally critical structural changes, such as corners, intersections, or new landmarks that semantic-only comparisons would otherwise overlook.

Classical approaches such as SIFT~\cite{lowe2004distinctive} or ORB~\cite{rublee2011orb} could, in principle, serve as lightweight alternatives, but they are highly sensitive to illumination changes and image noise; making them unsuitable for reliable, continuous gating. In Section~\ref{sec:Evaluation}, we quantify the benefits of this structurally-aware KFC approach compared to a semantic pooling baseline.


\begin{figure}[!t]
\centering
\includegraphics[width=1.0\linewidth]{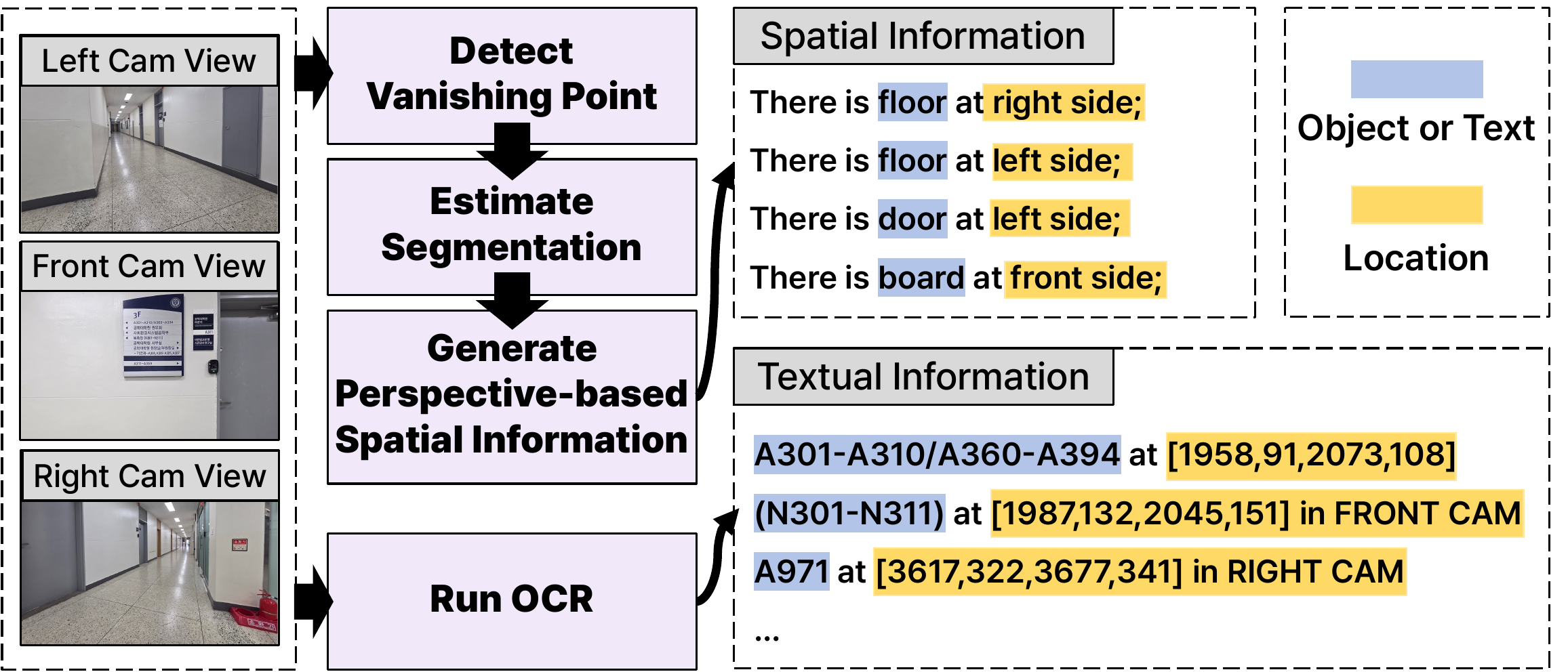}
\vspace{-4ex}
\caption{Spatial \& Textual information of the Scene}
\vspace{-2ex}
\label{fig:spatial_textual_figure}
\end{figure}

\subsubsection{Spatial and Textual Understanding}

Since \system{} executes its VLM entirely on on-device GPUs, the reasoning model must remain compact. To compensate for the limited spatial and textual capabilities of small VLMs, \system{} incorporates dedicated modules within the Feature Extractor for capturing spatial structure and textual cues from the scene.

For spatial understanding, input video frames pass the segmentation module (SegFormer-b0~\cite{xie2021segformer}) to gather 2D semantic masks, which \system{} enhances using a perspective-based heuristic. The system detects the primary vanishing point (VP)~\cite{lu20132}, typically aligned with the geometric center of hallways, and uses it to infer a coarse 3D layout. By projecting rays toward the VP, the scene is partitioned into navigational zones (``front,'' ``left,'' ``right,'' ``down'') and distinguished into ``immediate’’ and ``distant’’ regions, as in Figure~\ref{fig:spatial_textual_figure}.

For textual understanding, \system{} employs OCR (docTR~\cite{mindee2021doctr}) to extract symbolic cues such as room numbers and signs, which often provide critical context for indoor navigation.

While qualitative, this spatial–textual decomposition yields structured scene descriptions (e.g., ``A wall is on the front side’’) that are combined with OCR results to form the unified textual inputs consumed by both the \textit{Vision-Condition Matching} and \textit{Additional Thinking} components~\cite{medhi24targetprompting, zhang2024llavar}.

%
%

\begin{figure}[!t]
\centering
\includegraphics[width=1.0\linewidth]{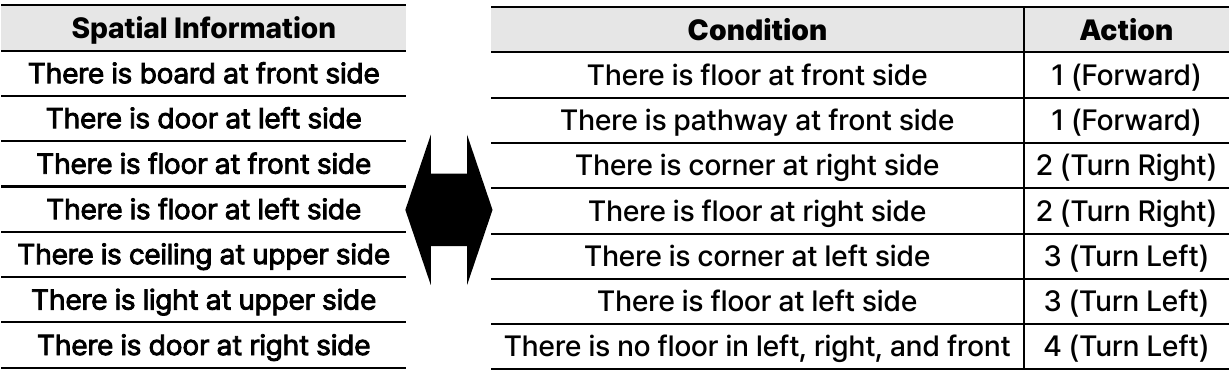}
\vspace{-4ex}
\caption{Mapping for spatial \& textual information to the Condition-to-Action table.}
\vspace{-2ex}
\label{fig:vision_condition_matching_table}
\end{figure}

\subsubsection{Condition Matching}

The \textit{Condition Matching} module in System One operates in two modes: (i) Destination Arrival Checking and (ii) Vision-Condition Matching

Destination Arrival Checking computes the vision-text similarity between the ``Goal Description'' and the visual embedding from the Vision Encoder, while simultaneously verifying the ``Destination Text'' extracted by the VLM during the Condition-to-Action generation. If visual similarity exceeds a predefined threshold (80\% in our implementation) and the ``Destination Text'' is detected by the OCR, the robot declares arrival and issues an \texttt{Idle} action.

\textbf{Vision-Condition Matching} is triggered only when KFC detects a structural change. As Figure~\ref{fig:vision_condition_matching_table} illustrates, the spatial and textual scene descriptions are compared against the Condition-to-Action table via cosine similarity. All entries above a heuristic threshold are aggregated into a \textit{UniqueActions} set, effectively serving as the decision-making process.

If this set contains exactly one action, it is executed by the finite-state \textit{Action Executor}. In our implementation, the Execution Module generates multiple Vanishing Point (VP) candidates representing possible movement directions and selects a single VP based on a designated rule. For CW movements, the module selects the rightmost VP; for CCW, it selects the leftmost VP; and for forward motions, it selects the oldest stable VP. Turn-back rotates the robot by $180^\circ \pm 20^\circ$ and selects a newly stable VP. If none or multiple actions are detected, the case is deemed ambiguous and escalated to System Two for \textit{Additional Thinking}.

\subsection{System Two: Deliberative Reasoning and Augmentation}

\textbf{System Two} serves as the reasoning core of \system{}, handling complex or ambiguous scenarios escalated from System One and augmenting the overall intelligence of the system. It is designed to enhance the cognitive and reasoning capability of any Vision-Language Model (VLM), particularly in spatially grounded tasks, and is based on two primary components:


\begin{enumerate}[leftmargin=*]
    \item \textbf{Condition-to-Action Generation} is an offline process to preload potential environmental conditions and corresponding robotic actions by allowing the VLM to ``speculatively inference'' possible scenes and plausible actions using a start location image and goal description prompt.
    \item \textbf{Additional Thinking:} A real-time reasoning module that enables deep scene understanding and final decision making in complex, uncertain, or unseen situations.
\end{enumerate}

\begin{figure}[!t]
\centering
\includegraphics[width=0.95\linewidth]{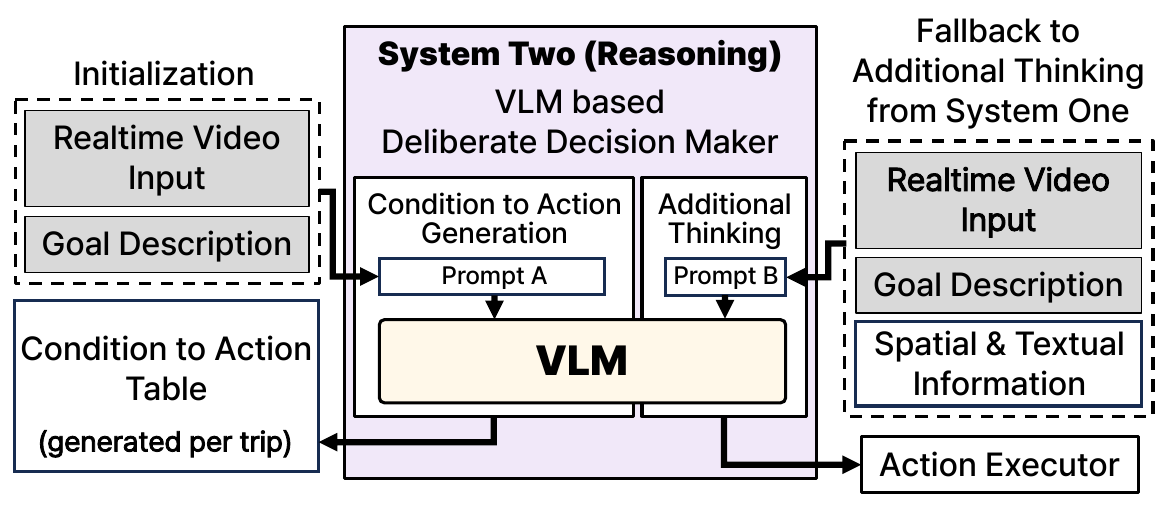}
\vspace{-3ex}
\caption{Detailed overview of System Two}
\vspace{-2ex}
\label{fig:system_two_overview}
\end{figure}

This dual mechanism allows the VLM to leverage augmented spatial \& textual information from System One, addressing the struggle of compact VLMs in perceiving geometric or topological structure and textual information. A high-level overview of System Two is presented in Figure~\ref{fig:system_two_overview}. Our implementation employs \texttt{Gemma3-4B} model as the VLM backbone; while we note that \system{} is model-agnostic and compatible with any VLM capable of multimodal reasoning.

\subsubsection{Offline Condition-to-Action Generation}

The \textit{Condition-to-Action Generation} precomputes a structured rule set used in System One. During initialization, the robot provides its current view image and goal description to the VLM via a goal-specific prompt. The prompt enforces a strict output format to ensure compatibility with downstream modules. Upon generation, the system validates and parses the output; if the format is incorrect, the VLM regenerates the rules until a valid table is produced; guaranteeing robustness and consistency in the rule base prior to deployment. Note that given the limited \textit{Action Set} of the robot, the combinations of Condition-Action pairs are limited; thus, limiting the scenario space and possibility of hallucination.

\subsubsection{Real-Time Additional Thinking and Spatial/Textual Augmentation}

When System One cannot determine a unique action, \textit{Additional Thinking} takes place. The input to the VLM in this case includes the goal description, the current scene image, and the spatial and textual information produced by System One. These spatial cues compensate for the limited geometric reasoning ability of compact VLMs.

A customized prompt constrains the VLM output to a fixed structure. To satisfy real-time constraints, we apply a maximum token budget and a logit processing~\cite{wolf20transformers} strategy, a technique used during the decoding phase to manipulate output probabilities; thereby guiding or restricting the model to generate specific tokens. In \system{}, as generation approaches $\sim$80\% of the token limit, logits for non-action tokens are progressively suppressed, guiding the model to terminate reasoning and output the final \textit{Action}.

If the VLM still fails to produce an action, the system restores the last key–value (K–V) cache and resumes generation, applying logit processing again. This ensures action completion with minimal latency while preserving the fidelity of VLM-based reasoning.

\subsection{Execution Module}

The \textbf{Execution Module} governs physical actuation and verifies whether actions are successfully executed. The \textit{Action Executor}, implemented as a finite-state machine, receives an \textit{Action} from either System One or System Two and transitions the robot into the corresponding state, executing the behavior defined in Figure~\ref{fig:system_overview}.

The module includes a real-time validation mechanism. In \system{}, this is handled by the \textbf{VP Tracker}, which estimates feasible directions by generating multiple Vanishing Point (VP) candidates. Upon issuing option, \system{} selects the VP aligned with the intended direction and steers the camera’s optical center toward it. Successful alignment confirms correct execution, enabling precise 360° maneuvering. 
We note that ultrasonic sensors on the robot platform trigger emergency stop in uncertain situations.


The Execution Module is extensible to diverse robot platforms and tasks. Developers may define custom states, actions, and success criteria, allowing the framework to be adapted to a wide range of physical AI systems.

\section{Evaluation}
\label{sec:Evaluation}

We now evaluate \system{} using an indoor navigation task to assess its accuracy, responsiveness, and robustness. Our experiments span five distinctive indoor environments and compare \system{} against VLM-only baselines to quantify the benefits of the proposed dual-process architecture.

\subsection{Experimental Settings}
\begin{figure}[!t]
\centering
\subcaptionbox{Full shot of robot}[0.45\linewidth]{
    \includegraphics[width=1.0\linewidth]{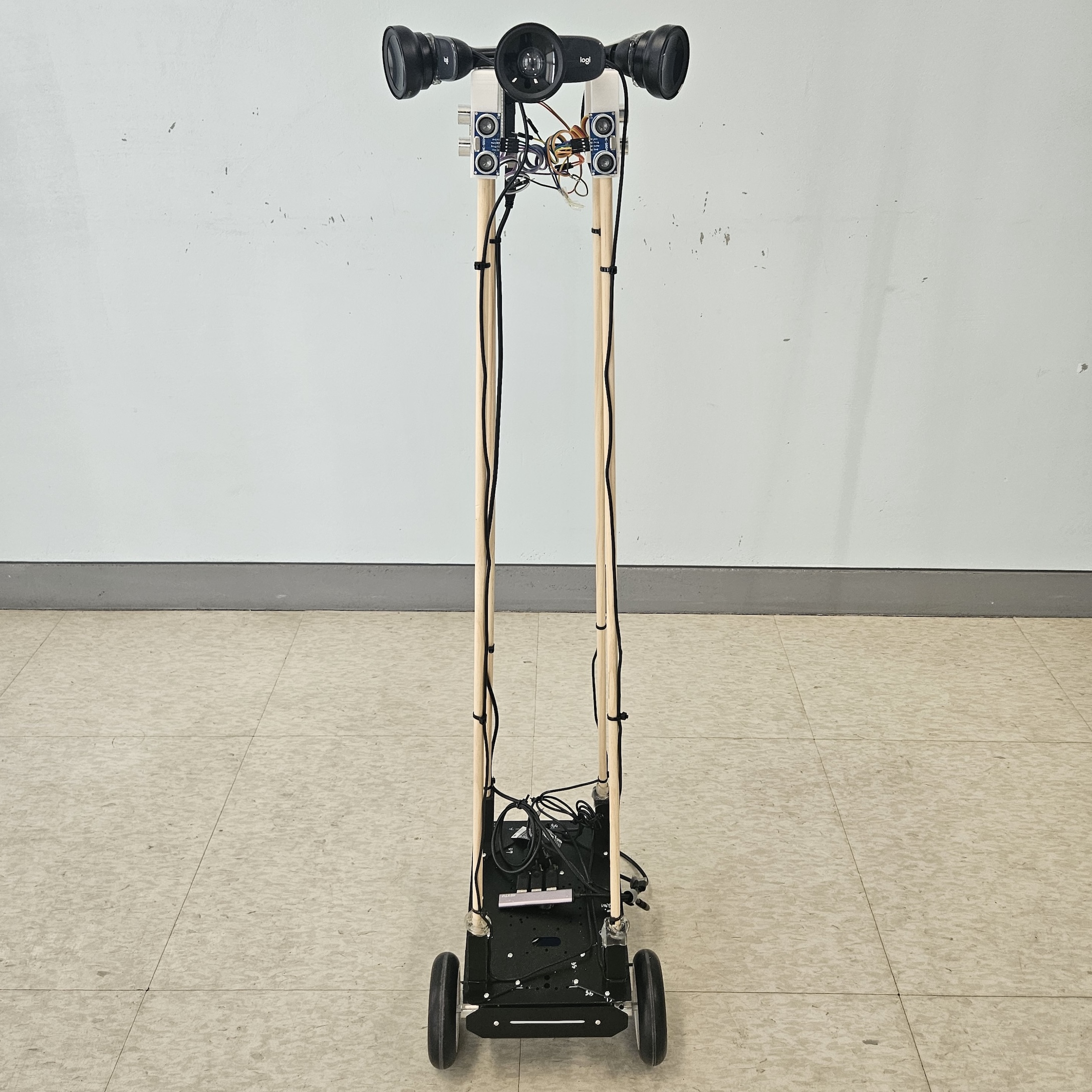} 
}
\hfill
\subcaptionbox{Head of Robot}[0.45\linewidth]{
    \includegraphics[width=1.0\linewidth]{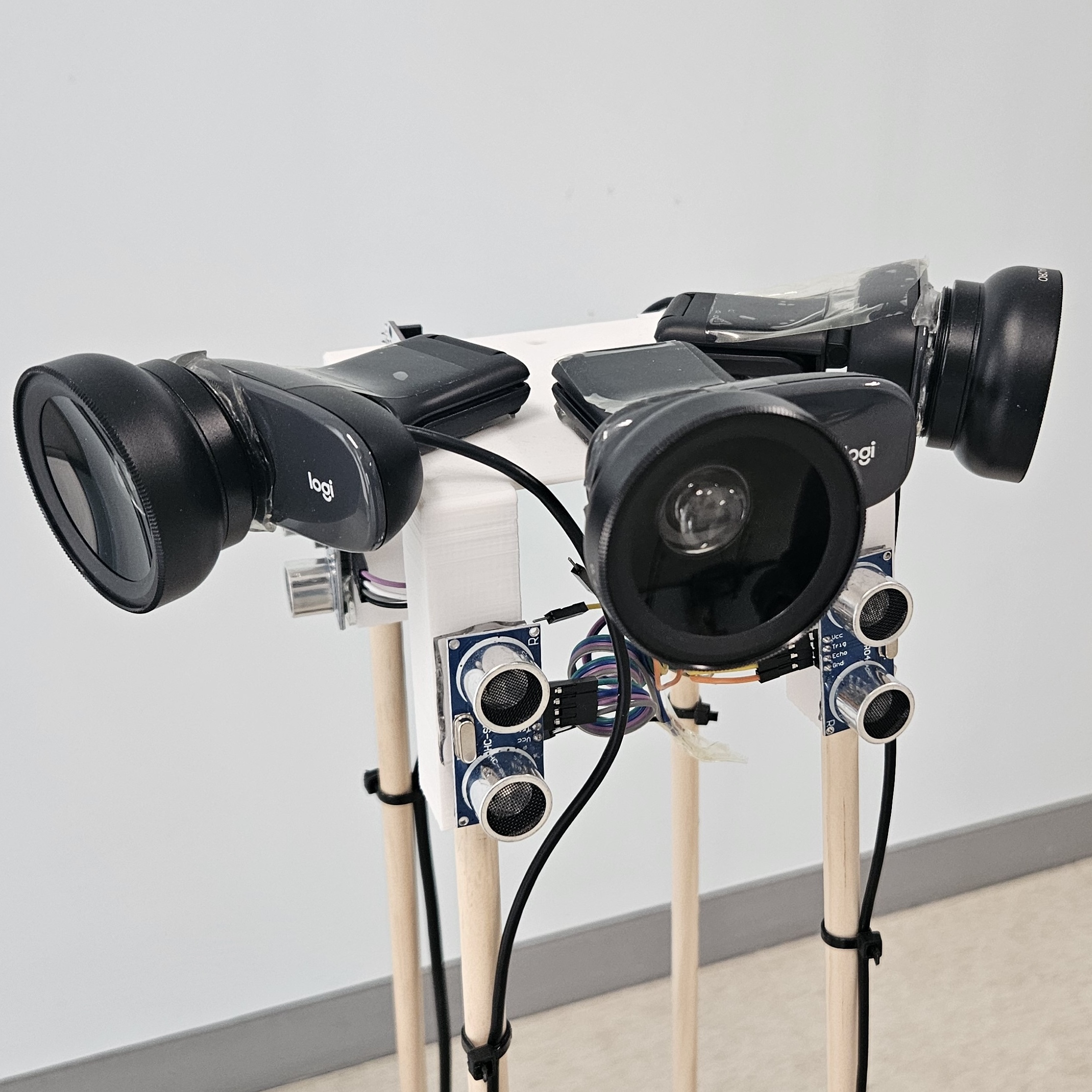}
}
\vspace{-3ex}
\caption{Test robot system}
\vspace{-2ex}
\label{fig:cerberus}
\end{figure}

\system{} was implemented on our testbed robot as in Figure~\ref{fig:cerberus}. The robot features three wide-angle cameras mounted on the upper frame and four wheels for omnidirectional movement. The camera layout is designed to capture human-targeted cues, such as signs, door plates, and room numbers, which are critical for indoor navigation. The robot runs on an \texttt{NVIDIA Jetson Orin NX}~\cite{nvidia2023orin} (MAXN mode) with 16\,GB of memory, supporting fully on-device operations. 

\begin{figure*}[!t]
\centering
\subcaptionbox{University 1}[0.19\linewidth]{
    \includegraphics[width=1.0\linewidth]{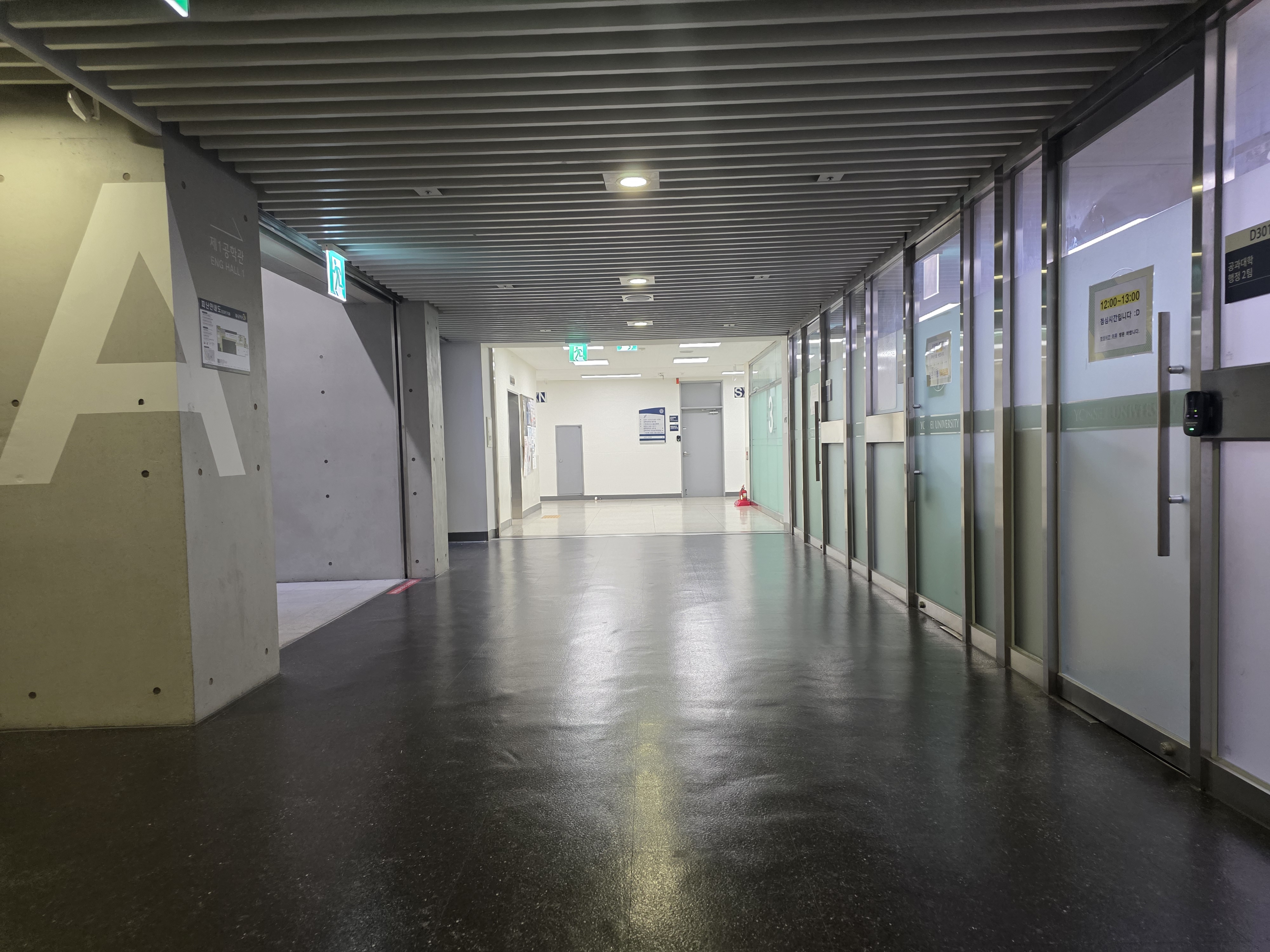} 
}
\hfill
\subcaptionbox{University 2}[0.19\linewidth]{
    \includegraphics[width=1.0\linewidth]{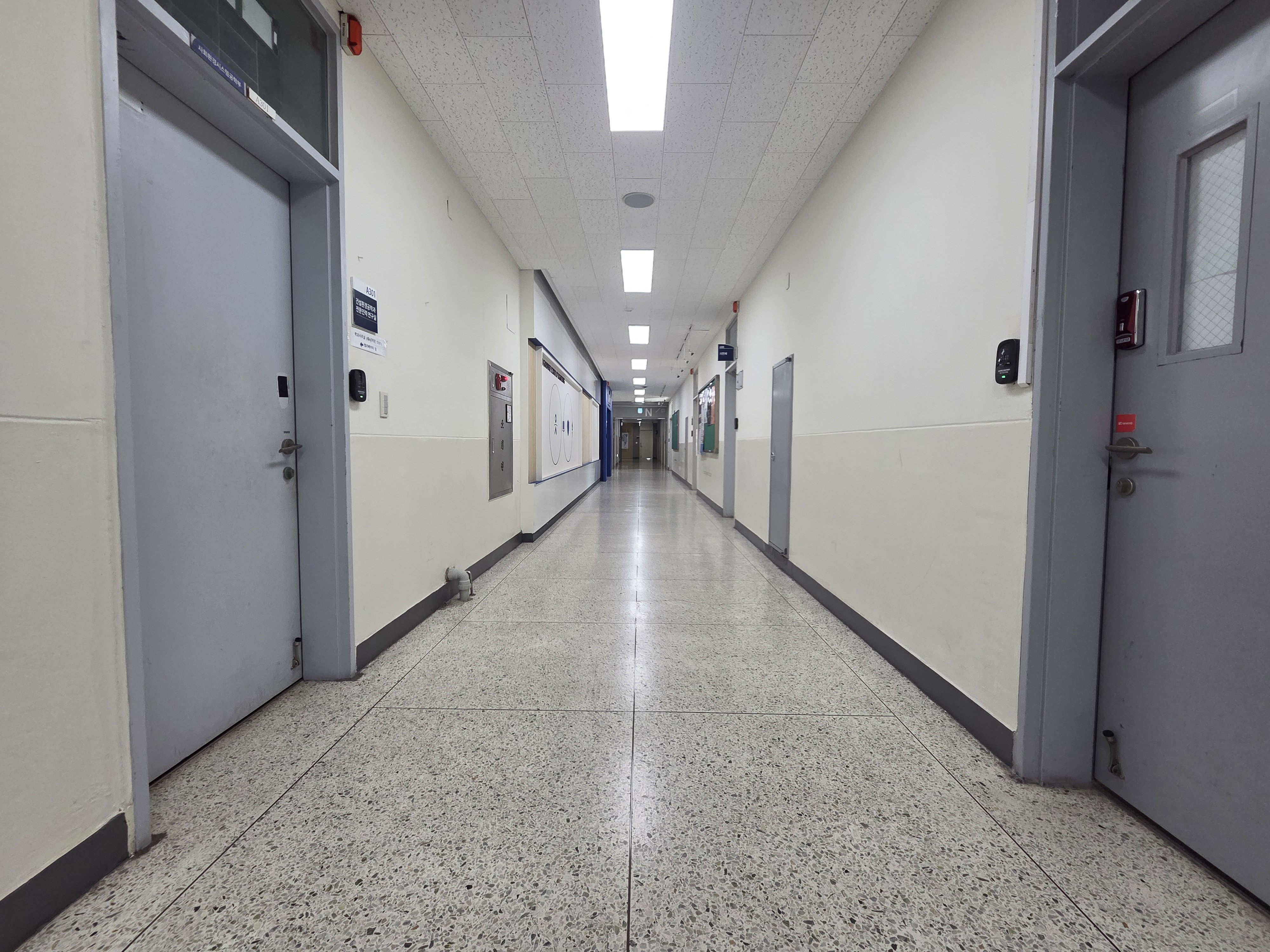} 
}
\hfill
\subcaptionbox{University 3}[0.19\linewidth]{
    \includegraphics[width=1.0\linewidth]{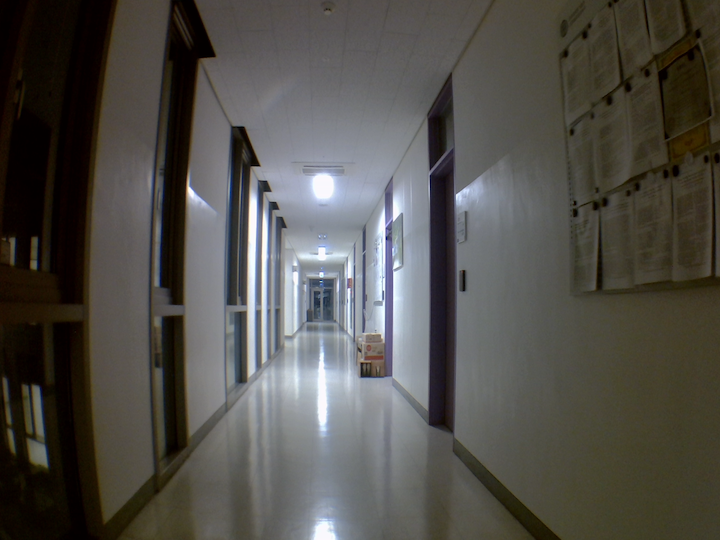} 
}
\hfill
\subcaptionbox{Office Complex}[0.19\linewidth]{
    \includegraphics[width=1.0\linewidth]{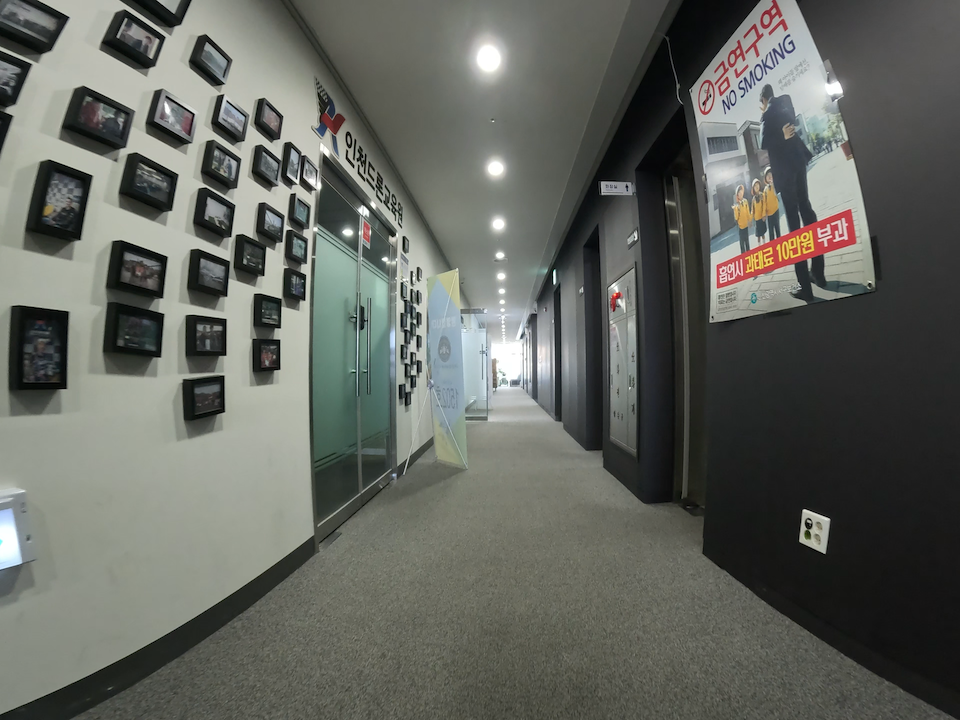} 
}
\hfill
\subcaptionbox{Residential}[0.19\linewidth]{
    \includegraphics[width=1.0\linewidth]{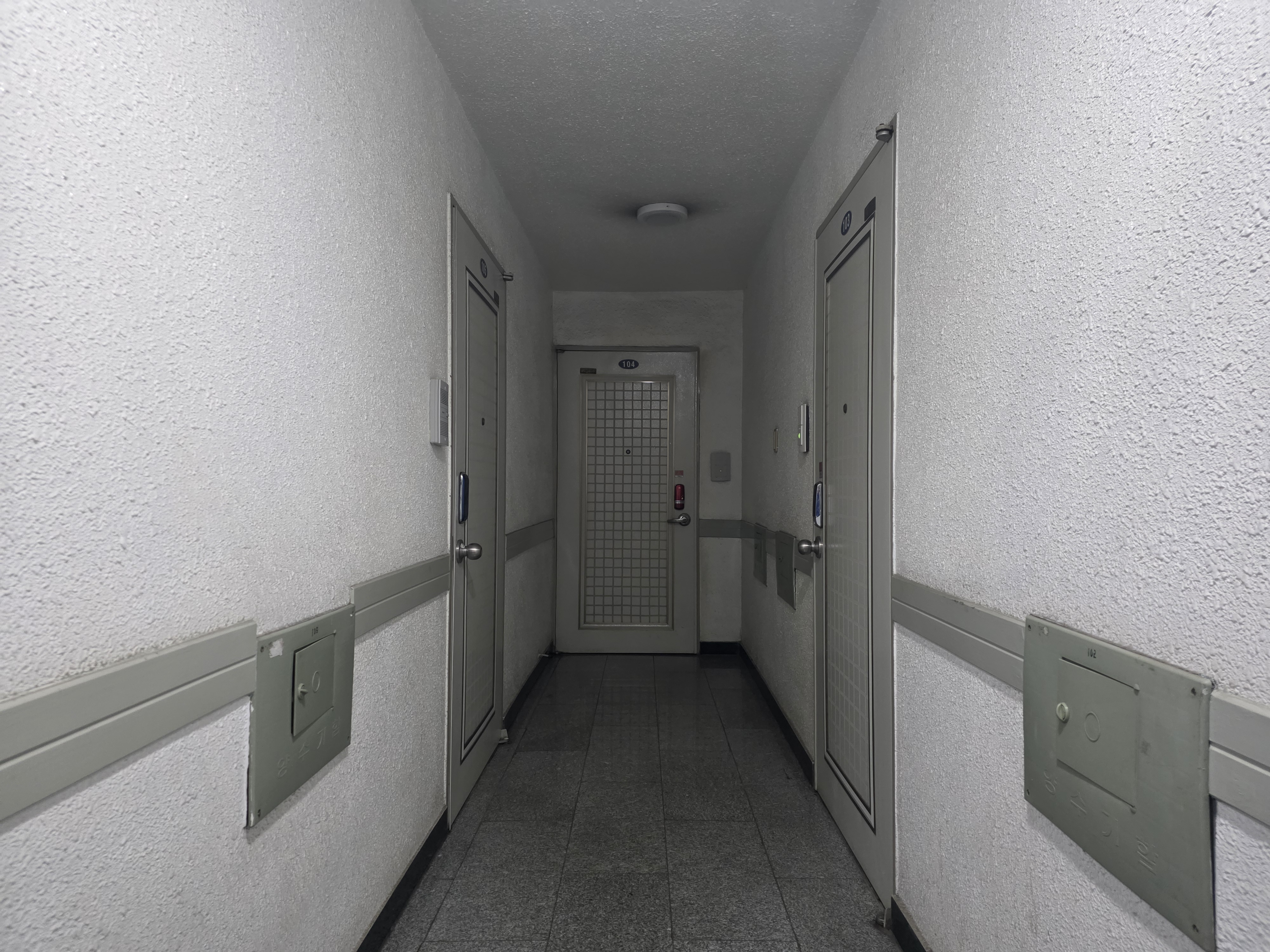} 
}
\vspace{-3ex}
\caption{Tested Environments}
\label{fig:test_env}
\end{figure*}

We evaluated \system{} across five real-world environments: three university buildings, one office complex, and one residential building (Figure~\ref{fig:test_env}). These sites were selected to reflect diverse structural layouts, lighting conditions, and representative scene challenges for typical indoor navigation tasks. ``University 1'' features extensive glass structures, while ``University 2'' and ``University 3'' share textureless walls but differ in structural layout. ``Office Complex'' contains dense visual clutter with irrelevant wall attachments alongside target signs, rigorously testing the VLM's ability to filter semantic noise. The ``Residential'' setting features minimal contrast and non-standard high signage, validating spatial augmentation robustness against visual ambiguity. We note that \system{}'s performance across these environments did not show significant differences; thus, we report results for a dataset that evenly combines data from all environments.

The number of target destinations in each environment was scaled according to its physical layout: three destinations for each university building, two for the office complex, and one for the residential building. For each scenario, we fixed the start location and varied the semantic goal description so that the robot traversed distinct branching points. We conducted 10 episodes per source–destination pair (total of 120 episodes or 2,455 meters of video footage) to ensure sufficient diversity and to rigorously evaluate whether \system{} can robustly perform human-context–aware decision making in heterogeneous environments.

To construct a human-reference baseline (i.e., ground truth), we collected action labels from five participants through an IRB-approved study. Each participant was given a floor map with marked start and goal positions and was provided sequential images captured by the robot every 50\,cm along each path. For each image, participants selected the action they found most appropriate from \{\textit{Forward}, \textit{CW($N^\circ$)}, \textit{CCW($N^\circ$)}, \textit{Turn Back}\}. For inconsistent responses, the discrepancy was discussed and converged to a single consensus action. This dataset serves as ground truth for assessing how \system{}’s decisions align with expected navigation behavior by mapping the most relevant labeled image with the robot input.

\subsection{Overall Decision-Making Performance}

We first evaluate \system{} end-to-end (i.e., the combined operation of System One, System Two, and all supporting modules) across all test environments. Each navigation task was specified semantically (e.g., ``go to room 306''), and the robot was required to reach the target location from its initial position. Due to hardware constraints, all evaluations were performed within a single floor of each building. Decision accuracy was computed by comparing \system{}'s navigational choice at each frame against the human-labeled baseline.

\begin{figure}[!t]
\centering
\subcaptionbox{Decision accuracy and latency}[0.65\linewidth]{
    \includegraphics[width=1.0\linewidth]{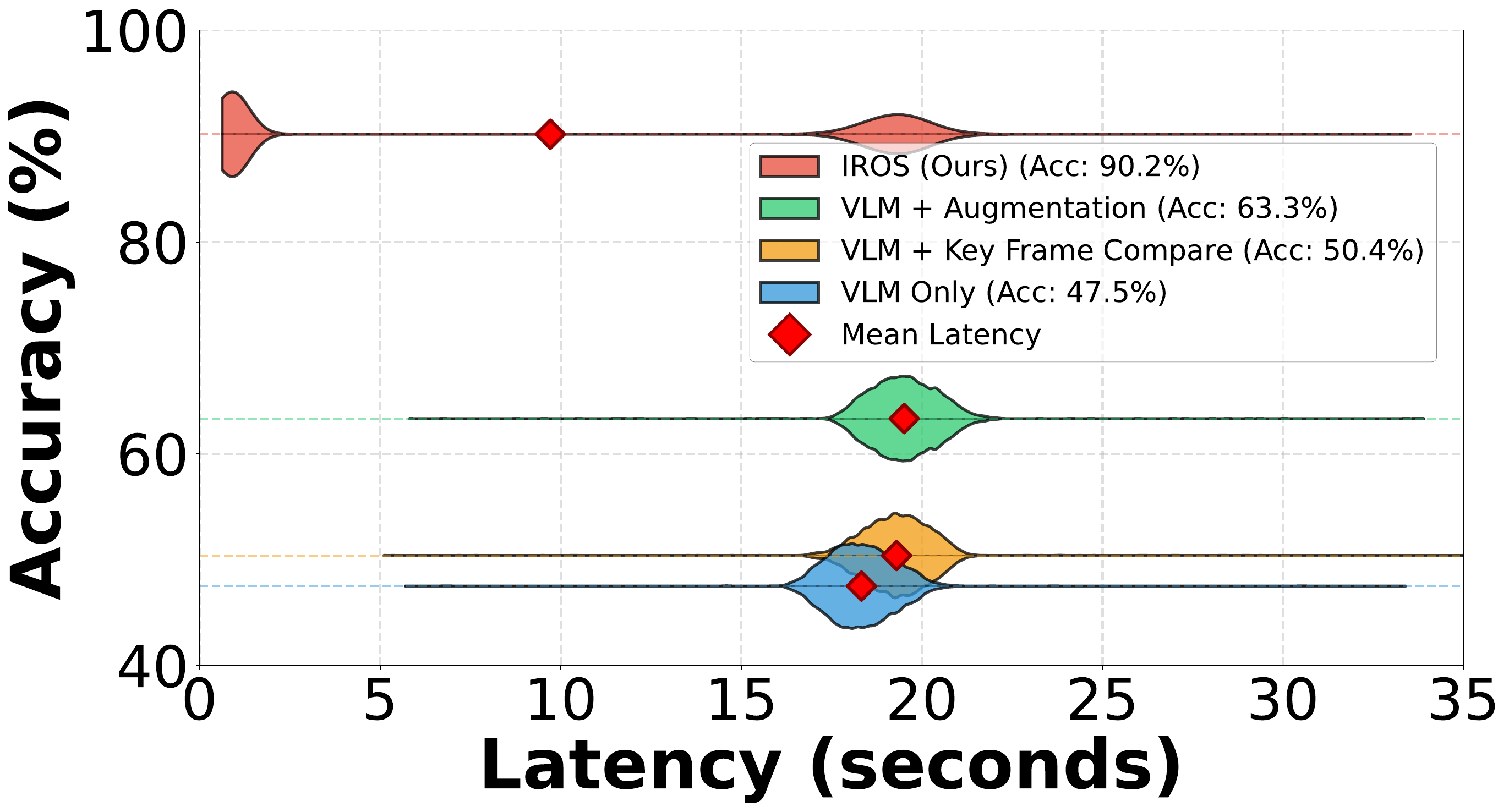} 
}
\hfill
\subcaptionbox{Per-system occupancy}[0.3\linewidth]{
    \includegraphics[width=1.0\linewidth]{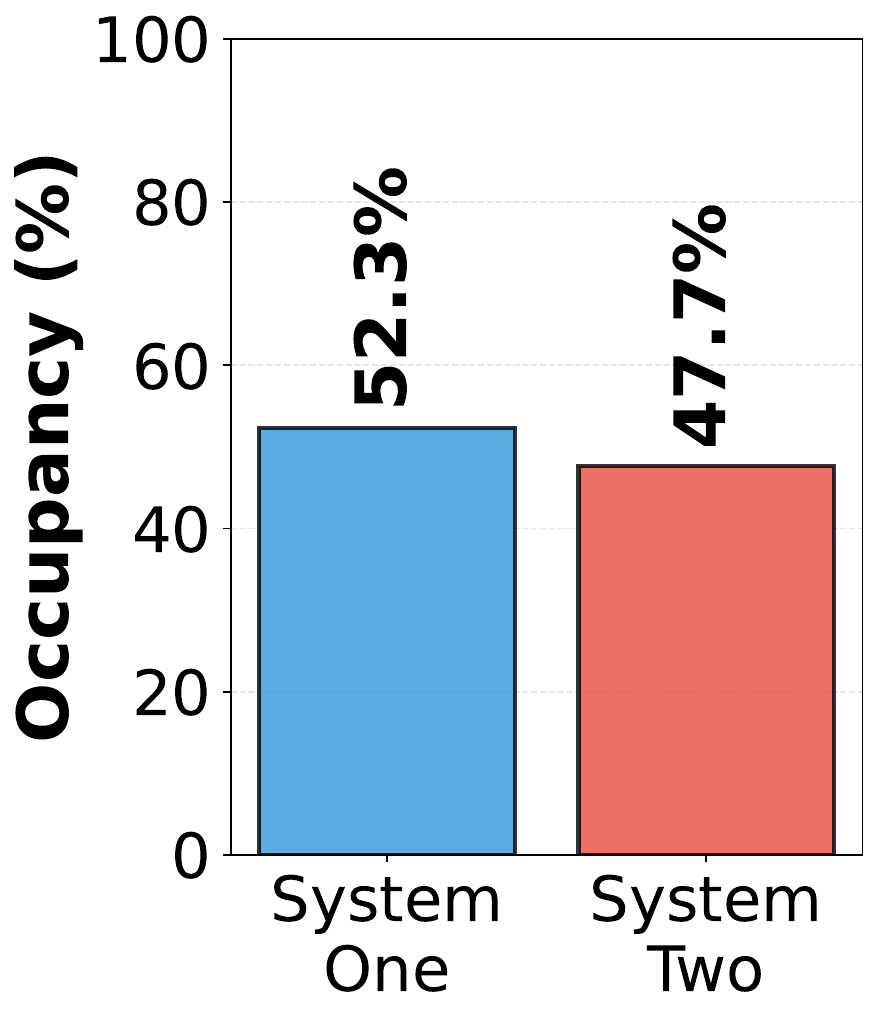}
}
\vspace{-3ex}
\caption{Overall decision making performance.}
\label{fig:overall_performance}
\end{figure}

As shown in the latency–accuracy plots of Figure~\ref{fig:overall_performance}(a), \system{} achieves high decision accuracy while operating at roughly half the latency of a VLM-only baseline (Gemma3 4B). We also note that, \system{} demonstrated a significantly higher end-to-end complete episode success rate of 67.5\% (11.5$\times$ than 5.83\% for VLM only). When observing the latency distribution, we notice that two distinct clusters exist for \system{}; one near $<$1\,s and another near 20\,s. This bimodal pattern arises from the dual-process architecture as decisions made by System One incur low latency, whereas decisions handled by System Two exhibit latency comparable to other VLM-based approaches. The occupancy (i.e., usage) distribution between Systems One and Two shown in Figure~\ref{fig:overall_performance}(b) corroborates this behavior, indicating that System One was responsible for 53.6\% of all navigational decisions. This reduction in reliance on expensive VLM inference directly enables the improved latency performance observed in \system{}.

\subsection{Performance of System One}

System One handles fast, reflexive navigation decisions with minimal computational overhead, making it an essential part of \system{} to minimize its navigation latency. We evaluate its performance across all environments in terms of latency, decision accuracy, and turnover behavior.

\subsubsection{System One Latency}

\begin{figure}[!t]
\centering
\includegraphics[width=0.85\linewidth]{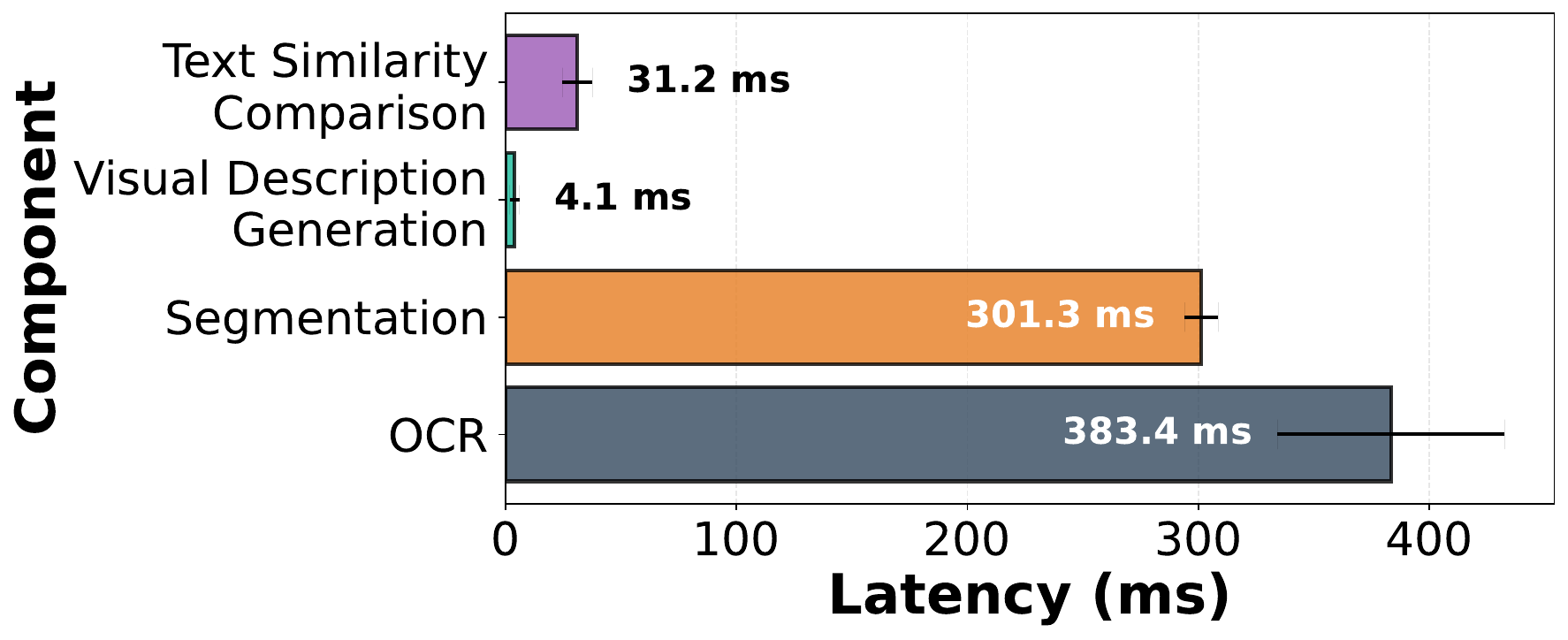}
\vspace{-2ex}
\caption{Latency breakdown for System One.}
\label{fig:latency_component}
\end{figure}

As Figure~\ref{fig:latency_component} shows, System One's end-to-end processing time remains between 0.7 and 0.9 seconds across all test environments, far faster than the 17–19 seconds typically required by System Two. This responsiveness is essential for safe operation in dynamic indoor settings. By breaking down the three components of System One, segmentation (301.3 ms), OCR(383.4 ms), text description generation (4.1 ms), and condition matching (31.2 ms), we notice that while the segmentation task takes up a large portion, the overall latency remains sufficiently low to preserve System One’s real-time advantage.

\subsubsection{Scene Augmentation via OCR \& Segmentation}

Extracting abstract visual features using a Vision Encoder (VE) is a widely adopted strategy for interpreting scene-level cues. The VE serves as a cross-modal interface, aligning visual features with textual embeddings for efficient visual-language interaction. Prior work such as FSR-VLN~\cite{zhou25fsrvlnfastslowreasoning} has demonstrated the utility of this approach through CLIP-based fast matching for contextual scene understanding in navigation.

%

However, VE-derived embeddings alone are insufficient for interpreting \emph{structural cues} critical for navigation. To validate this limitation, we constructed nine ``Condition-to-Action'' text templates by combining three structural terms (hallway, floor, corridor) with three spatial cues (front, left, right) (e.g., ``there is a hallway on the front side.''). Using these templates, we performed Vision-Condition Matching by computing cosine similarity between VE outputs and each textual condition. We evaluated three representative encoders: CLIP ViT-B/32 \cite{radford2021learning}, SigLIP-base-patch16-224 \cite{zhai2023sigmoid}, and BLIP2-OPT-2.7B \cite{li2023blip} (paired with \texttt{all-MiniLM-L6-v2}~\cite{wang20minilm} for text comparison). Since BLIP-2 exceeds Jetson Orin NX memory capacity, its experiments were done on an RTX 5060 server.

Thresholds for action selection were calibrated using a 20\% held-out validation set, sweeping the similarity threshold from 100\% to 0\% to identify the value that maximized precision. This mirrors the matching logic used in System One. Ground-truth labels indicated whether a turnover was required or, if not, which navigational action was correct.

\begin{figure}[!t]
\centering
\subcaptionbox{Accuracy comparison}[0.49\linewidth]{
    \includegraphics[width=1.0\linewidth]{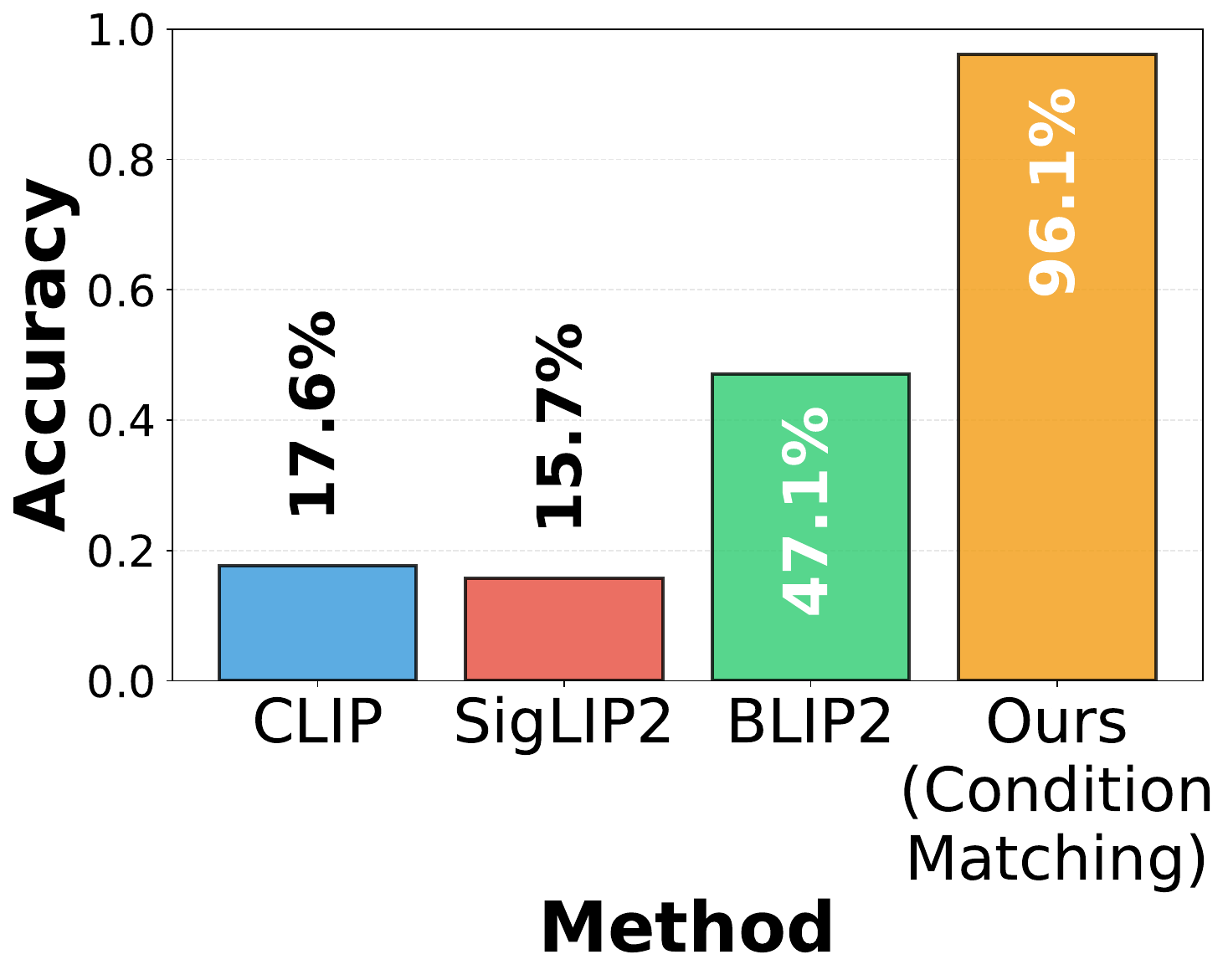} 
}
\hfill
\subcaptionbox{Latency comparison}[0.49\linewidth]{
    \includegraphics[width=1.0\linewidth]{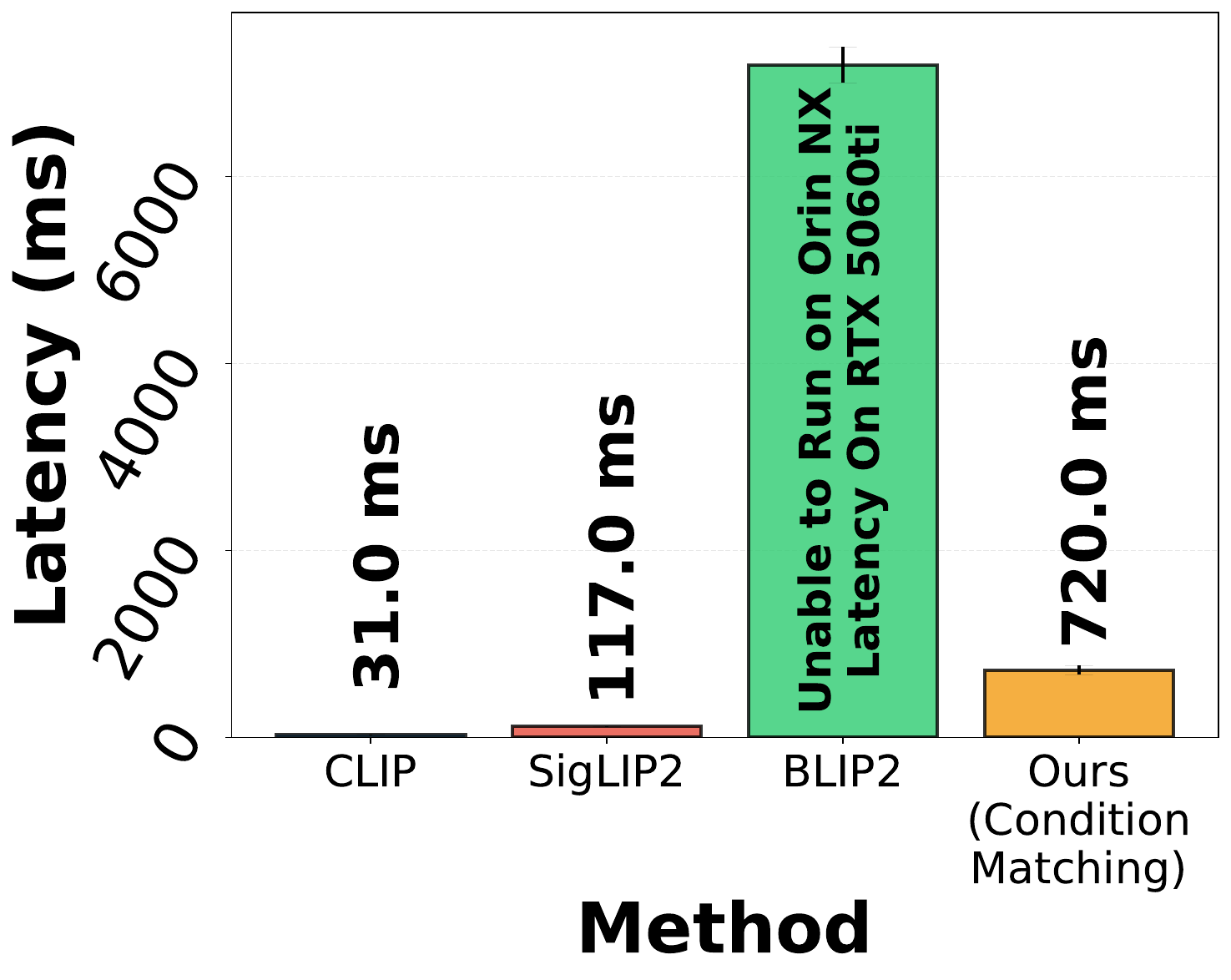}
}
\vspace{-2ex}
\caption{Performance comparison of VE Based \& Segmentation based Vision-Condition Matching.}
\vspace{-3ex}
\label{fig:ve_seg_compare}
\end{figure}

Results in Figure~\ref{fig:ve_seg_compare}(a) show that, standard VE-based similarity methods plateau below 50\% accuracy, which is far too low for reliable navigation. In contrast, our segmentation- and OCR-augmented Condition Matching method achieves 87.5\% accuracy, demonstrating its ability to capture structural cues that conventional VEs miss. Moreover, Figure~\ref{fig:ve_seg_compare}(b) shows that this improvement is achieved with sub-1-second latency, enabling structurally grounded scene understanding without sacrificing real-time performance.

\subsubsection{System One Decision Accuracy}

Next, we evaluate System One’s decision accuracy based on its ability to correctly identify safe, traversable paths without invoking higher-level reasoning. Condition Matching on Figure~\ref{fig:ve_seg_compare}(a) presents the per-frame accuracy of System One for the scenes (frame samples) that System One decided that it can take care of without passing it on to System Two. These results support two observations. First, it shows that System One achieves high accuracy when it determines that the current scene can be self-processed. Second, this result also implicitly suggests that System One accurately identifies the scenes that are suitable for fast processing. We will investigate this issue in depth using the next experiment.

\subsubsection{System One Turnover Performance}

\begin{figure}[!t]
\centering
\includegraphics[width=0.85\linewidth]{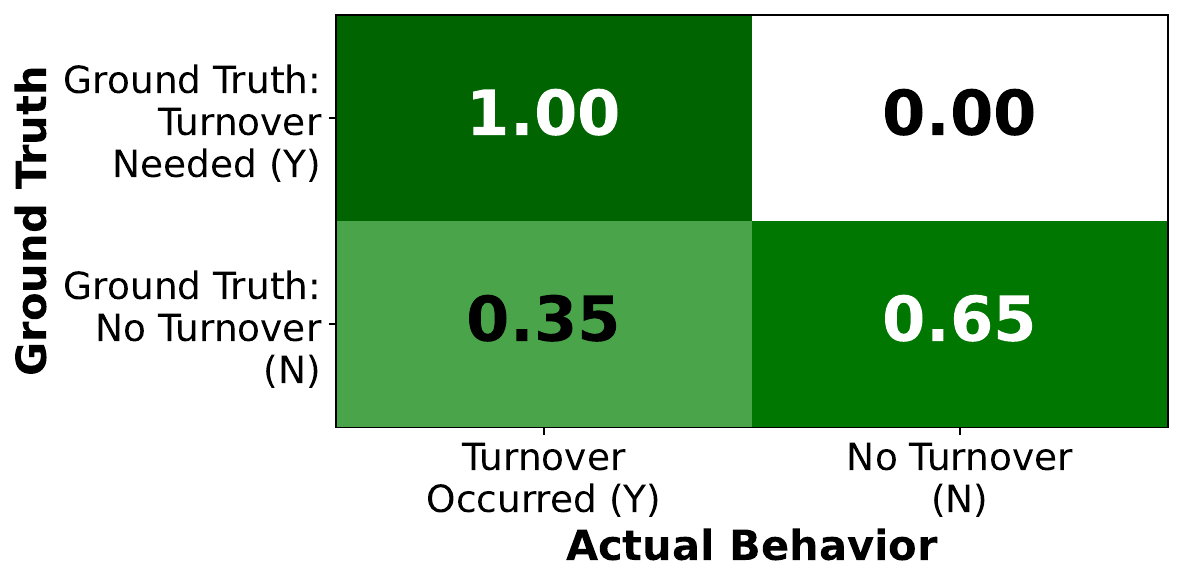}
\vspace{-3ex}
\caption{Turnover accuracy.}
\vspace{-3ex}
\label{fig:system_one_confusion_mat}
\end{figure}

A core component of \system{} is the \textit{turnover} mechanism, which determines when control should be delegated from System One to System Two. We evaluate the occurrence of this mechanism using a confusion matrix that compares human-labeled ground-truth turnover decisions with System One’s actual behavior in Figure~\ref{fig:system_one_confusion_mat}. By design, System One should handle only clear, unambiguous situations where a single action is evidently correct (e.g., continuing straight in a hallway), while System Two should assume control in more complex scenarios such as T-junctions, ambiguous intersections, or scenes with multiple plausible actions.

We report that \system{} made correct turnover decisions in $\sim$72\% of all cases (overall accuracy). More specifically, Figure~\ref{fig:system_one_confusion_mat} shows that System One achieves a 100\% recall rate for necessary turnovers, indicating it never fails to delegate control in critical situations. Despite the conservative behavior of deferring $\sim$35\% of manageable cases (False Positives), which limits \system{} from achieving maximum efficiency, our results show the robustness of \system{} by ensuring that potential misjudgments are resolved through VLM intervention rather than propagating into incorrect robot actions. 


\begin{figure}[!t]
\centering
\includegraphics[width=0.90\linewidth]{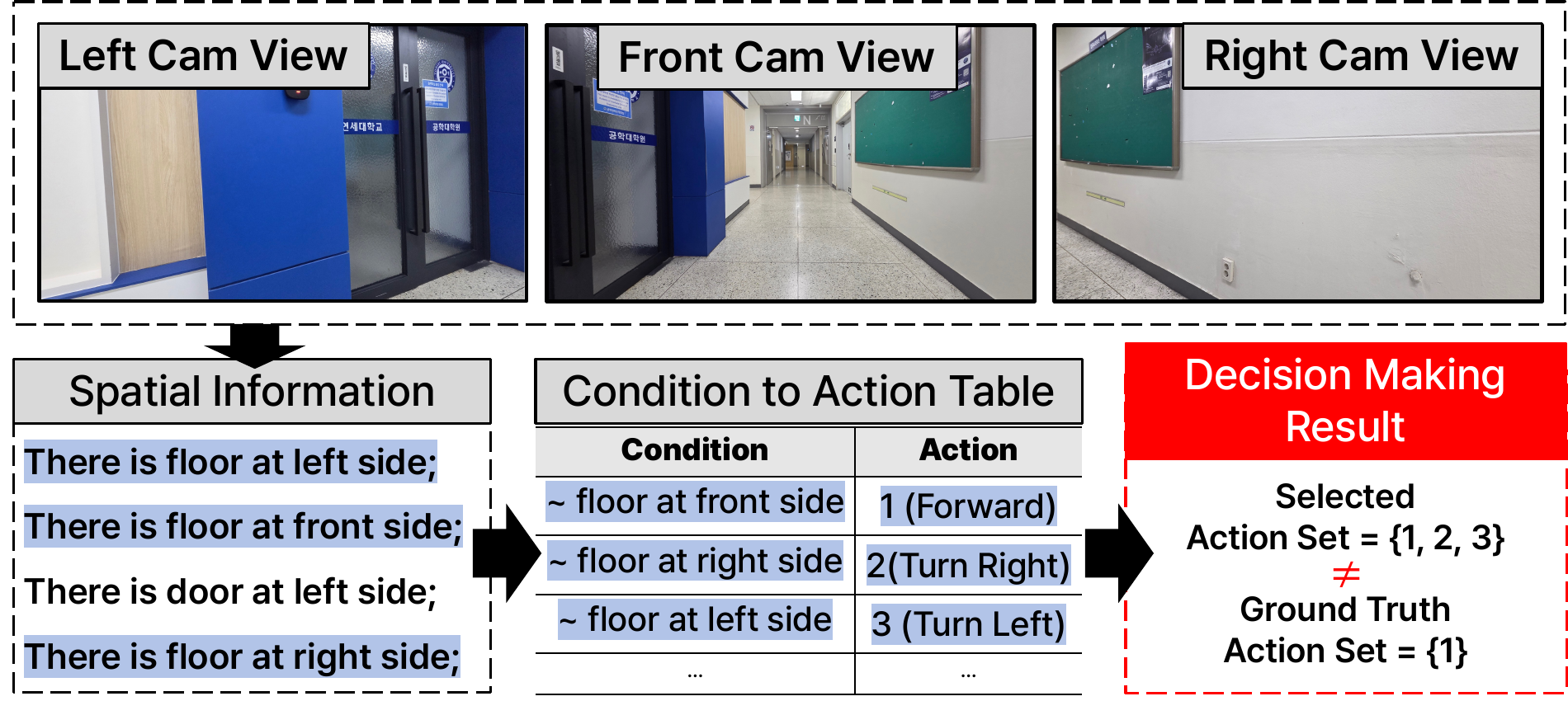}
\vspace{-2ex}
\caption{Qualitative analysis on turnover failure.}
\vspace{-3ex}
\label{fig:turnover_failure_analysis}
\end{figure}


To better understand the results in Figure~\ref{fig:system_one_confusion_mat}, we analyze two forms of undesirable turnover outcomes: (1) \emph{Turnover Occurring When Not Required (false positive case)} and (2) \emph{Turnover Not Occurring When Required (false negative case)}.

\begin{itemize}[leftmargin=*]

    \item \textbf{Turnover Occurred When Not Required (False Positive):} Figure~\ref{fig:turnover_failure_analysis} visualizes a sample decision pipeline where the system incorrectly selects multiple possible actions $\{1, 2, 3\}$ (Forward, CW, CCW) instead of the single ground truth $\{1\}$ (Forward), resulting in a False Positive case. Analysis reveals that the system erroneously identified the floor regions in the bottom-right corner of the Left Cam View and the bottom-left corner of the Right Cam View as navigable paths, leading it to perceive non-existent side passages. Consequently, false positives arise when slight expansions of segmentation masks into adjacent spatial regions are interpreted as structural changes. This hypersensitivity can cause System One to perceive non-existent junctions or forks, triggering unnecessary turnover events. While these turnovers introduce delays by invoking the VLM prematurely, they never lead to navigation failures. Instead, they function as a conservative safeguard that favors caution and correctness over speed.
    
    \item \textbf{Turnover Not Occurred When Required (False Negative):} We note that, our experiments observed a 0\% rate of this failure mode. This is attributed to our conservative spatial alignment strategy, which assigns spatial tags even when segmentation masks minimally overlap with their corresponding regions (e.g., front, left, right). This intentional bias causes System One to ``over-act,'' ensuring that any potentially ambiguous or high-risk scenario reliably escalates to System Two. As a result, the system consistently delegates complex or safety-critical decisions to the VLM, preventing missed turnovers entirely.

\end{itemize}


\subsection{Performance of System Two}

We now evaluate System Two, which handles complex reasoning tasks that require higher-level semantic interpretation and spatial understanding. We note that the Condition to Action Table is initially generated once per trip and took 25.25-32.89 sec for generating 6-9 condition-action pairs.

\subsubsection{Spatial Information Augmentation Performance}

Since \system{} runs entirely on an on-board GPU, it employs a compact VLM (4B-scale). A major challenge of this design is their limited ability to infer spatial structure from raw images. To mitigate this, \system{} augments the VLM’s input with structured spatial and textual cues generated by System One.

\begin{figure}[!t]
\centering
\includegraphics[width=0.85\linewidth]{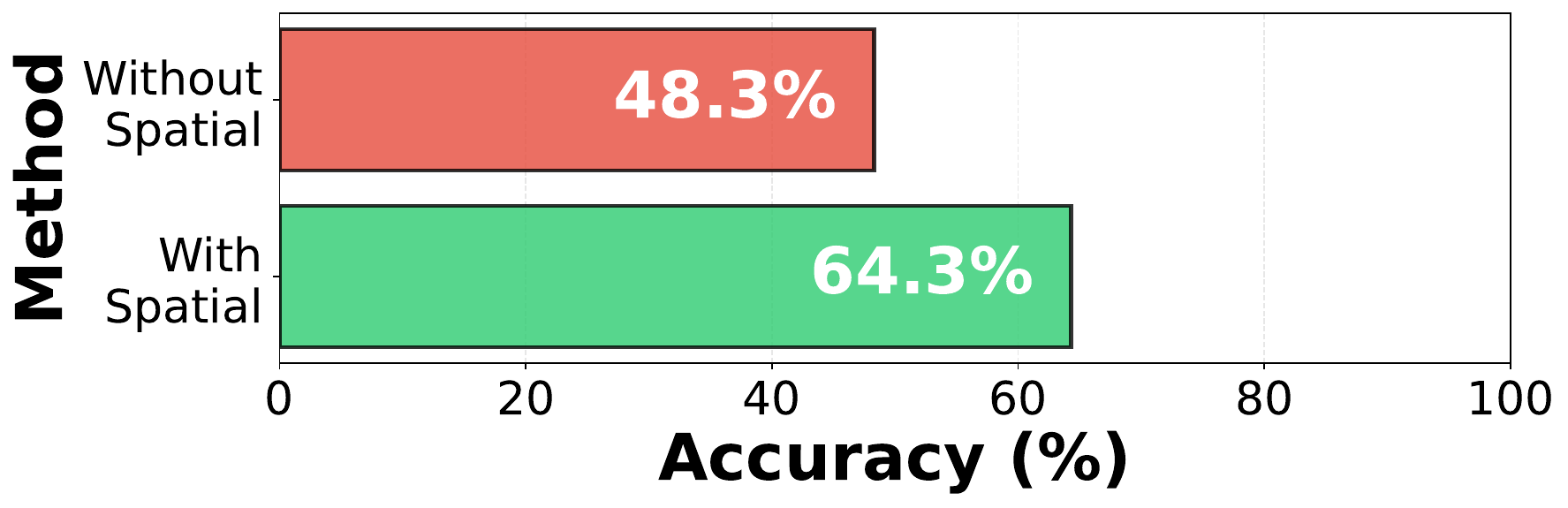}
\vspace{-2ex}
\caption{Decision accuracy comparison with and without Spatial \& Textual Information Augmentation.}
\label{fig:spatial_texual_augmentation}
\end{figure}

We compare the decision accuracy with and without the spatial information augmentation approach and present the results in Figure~\ref{fig:spatial_texual_augmentation}. As the results show, adding explicit spatial descriptions to the prompt noticeably improves the VLM’s ability to make correct navigational decisions.

\begin{figure}[!t]
\centering
\includegraphics[width=0.95\linewidth]{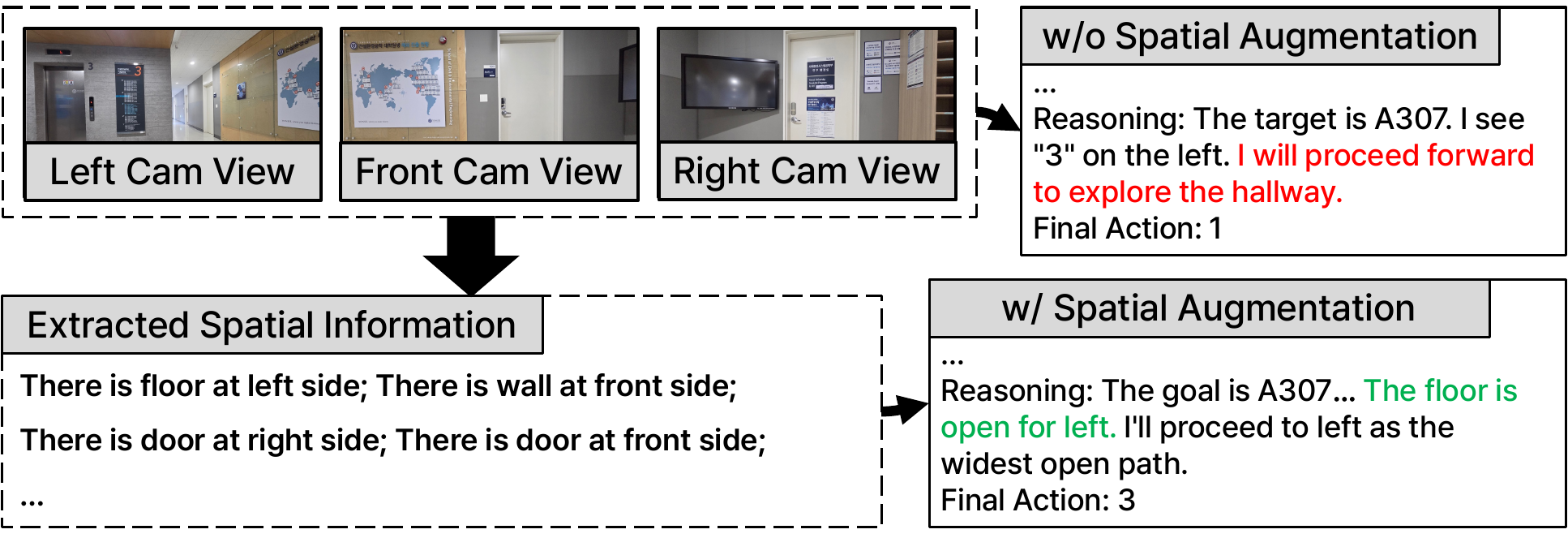}
\vspace{-2ex}
\caption{Impact of spatial information augmentation on Additional Thinking module performance.}
\vspace{-2ex}
\label{fig:augmentation_spatial_qualitative}
\end{figure}

\begin{figure}[!t]
\centering
\includegraphics[width=0.95\linewidth]{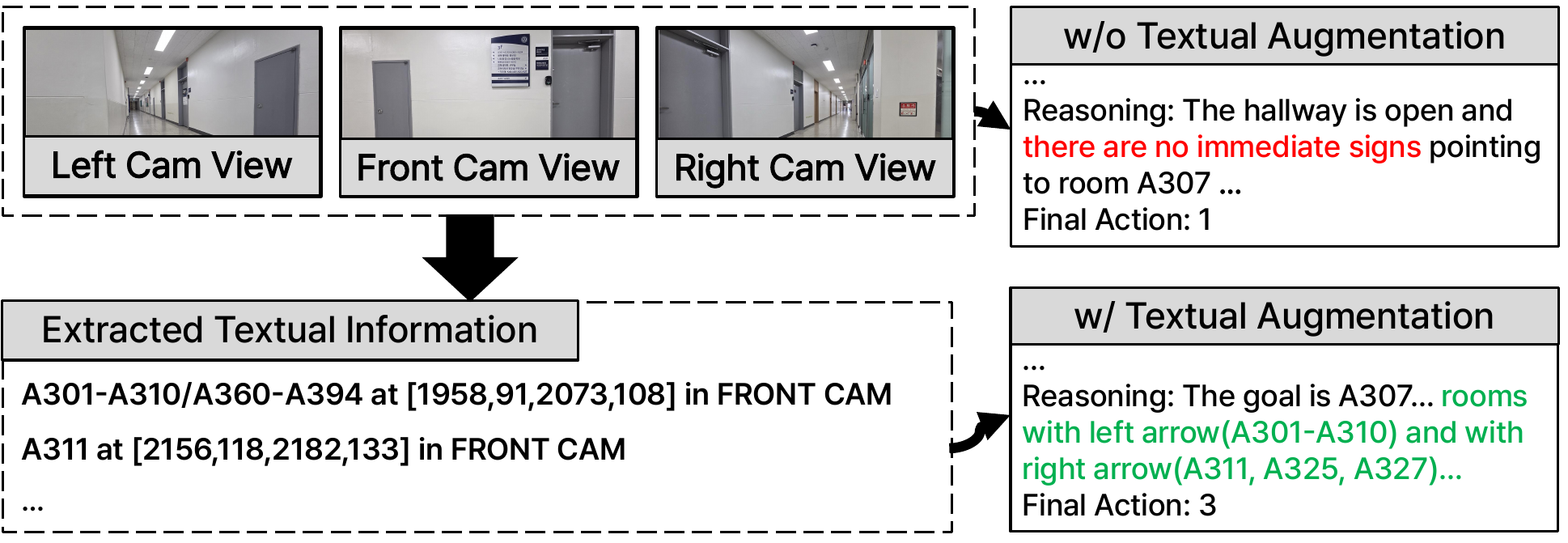}
\vspace{-2ex}
\caption{Impact of textual information augmentation on Additional Thinking module performance.}
\vspace{-2ex}
\label{fig:augmentation_textual_qualitative}
\end{figure}

Next, we present examples demonstrating how spatial and textual augmentation improves System Two’s reasoning. Figure~\ref{fig:augmentation_spatial_qualitative} illustrates a case where spatial augmentation enhances decision quality. When provided with the prompt ``There is a floor on the left,'' the VLM produces a more accurate scene interpretation and selects a correct action compared to the non-augmented case. Similarly, Figure~\ref{fig:augmentation_textual_qualitative} shows the effect of textual augmentation using the prompt ``A301–A310 text on front cam.'' Incorporating this cue enables the VLM to better interpret the environment and align its reasoning with the correct navigational intent. Together, these examples show how explicit cues from System One can guide System Two towards grounded, contextually consistent reasoning that reflects real-world spatial and semantic structures.

\subsection{Analysis on Navigation Efficiency}

\begin{figure}[!t]
    \centering
    \subcaptionbox{VLM only}[0.3\linewidth]{%
    \includegraphics[width=1.0\linewidth]{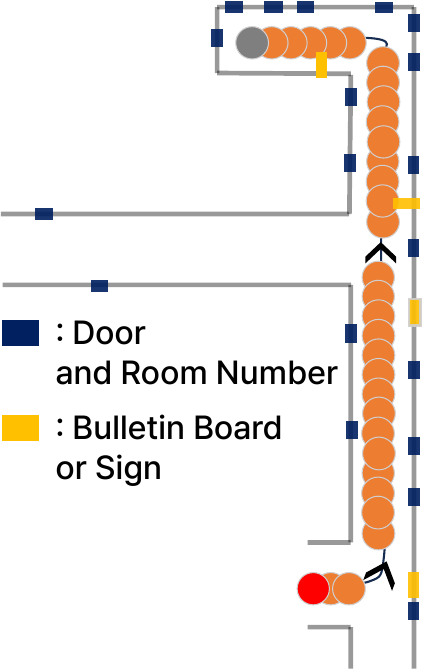}}
    \hfill
    \subcaptionbox{VLM + Vision Similarity (KFC)}[0.3\linewidth]{%
    \includegraphics[width=1.0\linewidth]{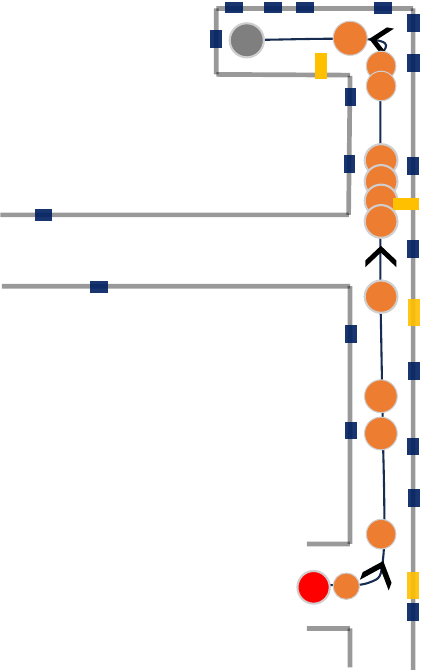}}
    \hfill
    \subcaptionbox{\system}[0.3\linewidth]{%
    \includegraphics[width=1.0\linewidth]{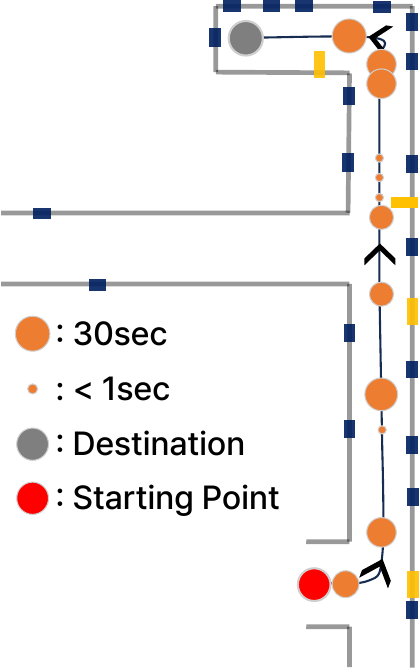}}
    \vspace{-3ex}
    \caption{Navigation decision latency and frequency for different system configurations.}
    \vspace{-2ex}
    \label{fig:route_analysis}
\end{figure}

Figure~\ref{fig:route_analysis} visualizes the navigation efficiency of different system configurations in reaching the target destination (gray dot) from the initial position (red dot). The size and location of each intermediate circle represents the inference latency and decision making locations, respectively.

We plot in Figure~\ref{fig:route_analysis}(a) a baseline configuration when only using a VLM. Here, the robot spends substantial time repeatedly invoking the VLM, as inference occurs for each captured image. In contrast, Figure~\ref{fig:route_analysis}(b) illustrates the addition of KFC as a preprocessing stage, which substantially reduces VLM activation frequency. This indicates that the baseline model performs redundant reasoning cycles, while the KFC-gated approach minimizes unnecessary computation and improves overall navigation efficiency.

The result for \system{} is illustrated in Figure~\ref{fig:route_analysis}(c). As shown, \system{} achieves a clear reduction in both the number and duration of decision-making instances. This improvement stems from its dual-process architecture with the KFC selectively gating VLM invocation, while Condition–Action Matching accelerates inference whenever possible. In this configuration, VLM execution functions only as a fallback for unresolved cases. In the example, System One handled 4 of 14 total decisions, accounting for approximately 29\% of the overall decision-making load.

\begin{figure}[!t]
\centering
\includegraphics[width=0.85\linewidth]{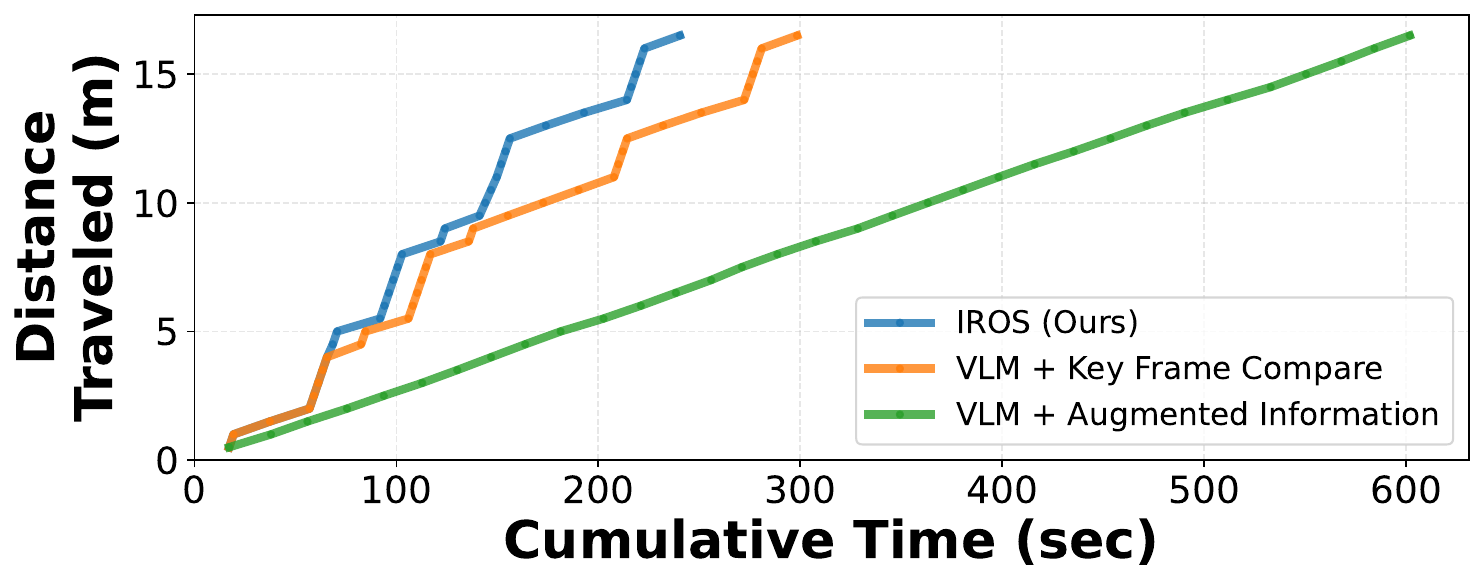}
\vspace{-2ex}
\caption{Time-travel distance for 16.5 m path.}
\vspace{-2ex}
\label{fig:time_n_travel_dist}
\end{figure}

Finally, Figure~\ref{fig:time_n_travel_dist} presents the results from Figure~\ref{fig:route_analysis} in terms of the time-to-travel distance ratio. Here, we add a case where the VLM is accompanied with the spacial information augmentation given that the relying solely on the VLM failed to reach the destination even after exceeding 600 secs of operation. This suggests that a naive VLM architecture is incapable of completing time-critical missions when deployed on-device due to its limited model capacity. In contrast, \system{} reached the destination (16.5 m away) in 240 secs, achieving a time saving of approximately 60\% compared to the VLM with the augmented information approach. This demonstrates that our proposed system operates efficiently and is applicable for time-critical tasks, while also minimizing latency-induced degradation of the user experience.

\subsection{Ablation Study}

Next we conducted a series of ablation studies to justify our design choices in terms of model size, effort, and thresholds.

\subsubsection{Key Frame Compare Threshold Tradeoff}

We first examine how the similarity threshold used in the Key Frame Compare (KFC) process affects \system{}’s scheduling behavior. Specifically, this threshold controls the KFC gate sensitivity, determining when inference should be triggered.

\begin{figure}[t]
\centering
\subcaptionbox{Sensitivity:60\%}[0.3\linewidth]{%
    \includegraphics[width=1.0\linewidth]{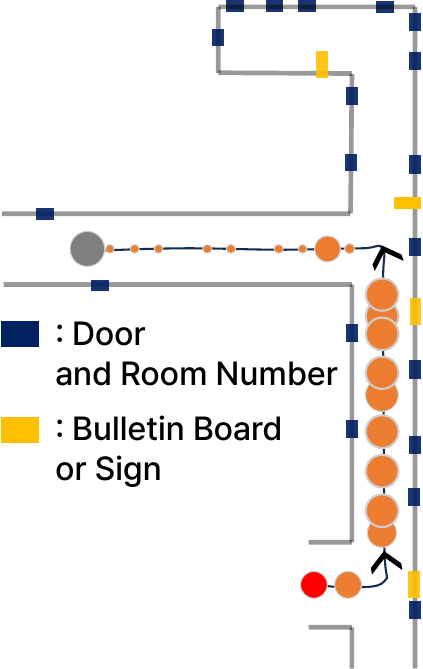}}
\hfill
\subcaptionbox{Sensitivity:45\%}[0.3\linewidth]{%
    \includegraphics[width=1.0\linewidth]{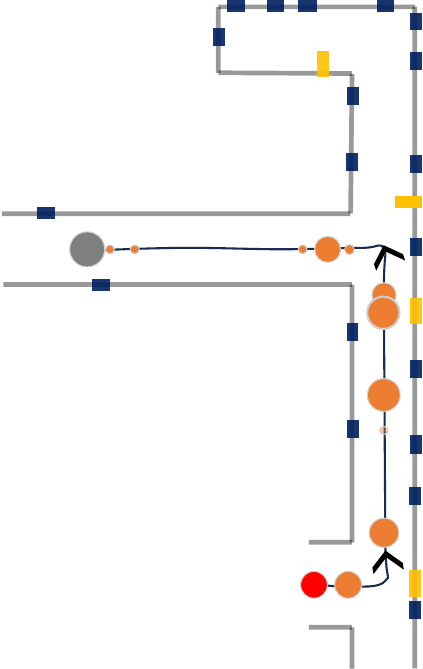}}
\hfill
\subcaptionbox{Sensitivity:30\%}[0.3\linewidth]{%
    \includegraphics[width=1.0\linewidth]{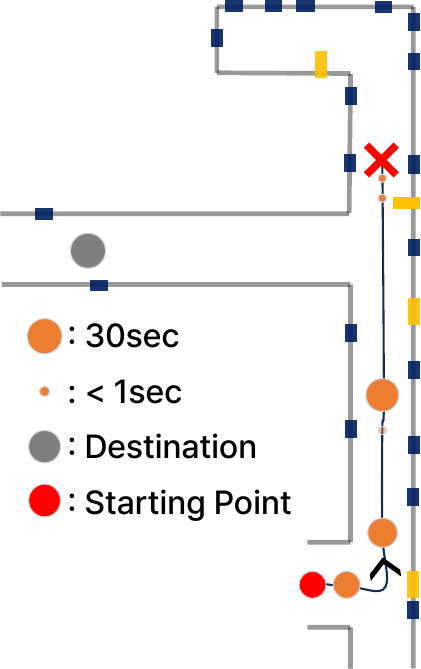}}
\vspace{-2ex}
\caption{Navigation decision latency and frequency for varying key frame compare similarity thresholds.}
\vspace{-2ex}
\label{fig:vis_sim_route_behavior}
\end{figure}


As Figure~\ref{fig:vis_sim_route_behavior}(a) illustrates, setting the threshold too high (60\%) causes the system to treat minor visual variations as structural changes, resulting in excessive inference calls. While this guarantees frequent decision-making opportunities, it leads to redundant computation. For instance, in a straight hallway, a 60\% threshold produced 21 inference requests; lowering the threshold to 45\% reduced this number to 13 (Figure~\ref{fig:vis_sim_route_behavior}(b)). While a few redundant calls still occur, the overall inference load is substantially reduced.

Conversely, an overly low threshold makes the system insensitive to meaningful structural changes. As illustrated in Figure~\ref{fig:vis_sim_route_behavior}(c), the robot failed to trigger inference at a left-turn junction and continued straight. The red ``x'' marks the time when \textit{Additional Thinking} was later invoked, but the VLM, now too far past the junction, recommended continuing straight rather than turning back, forcing us to terminate the trial. Despite these failure modes at extreme settings, a 45\% threshold demonstrated consistently stable performance across all five environments. This indicates that \system{} maintains robust behavior under reasonable parameter choices and is resilient to environmental variability.

\subsubsection{VLM Accuracy–Latency Tradeoff}

\begin{figure}[!t]
\centering
\includegraphics[width=0.85\linewidth]{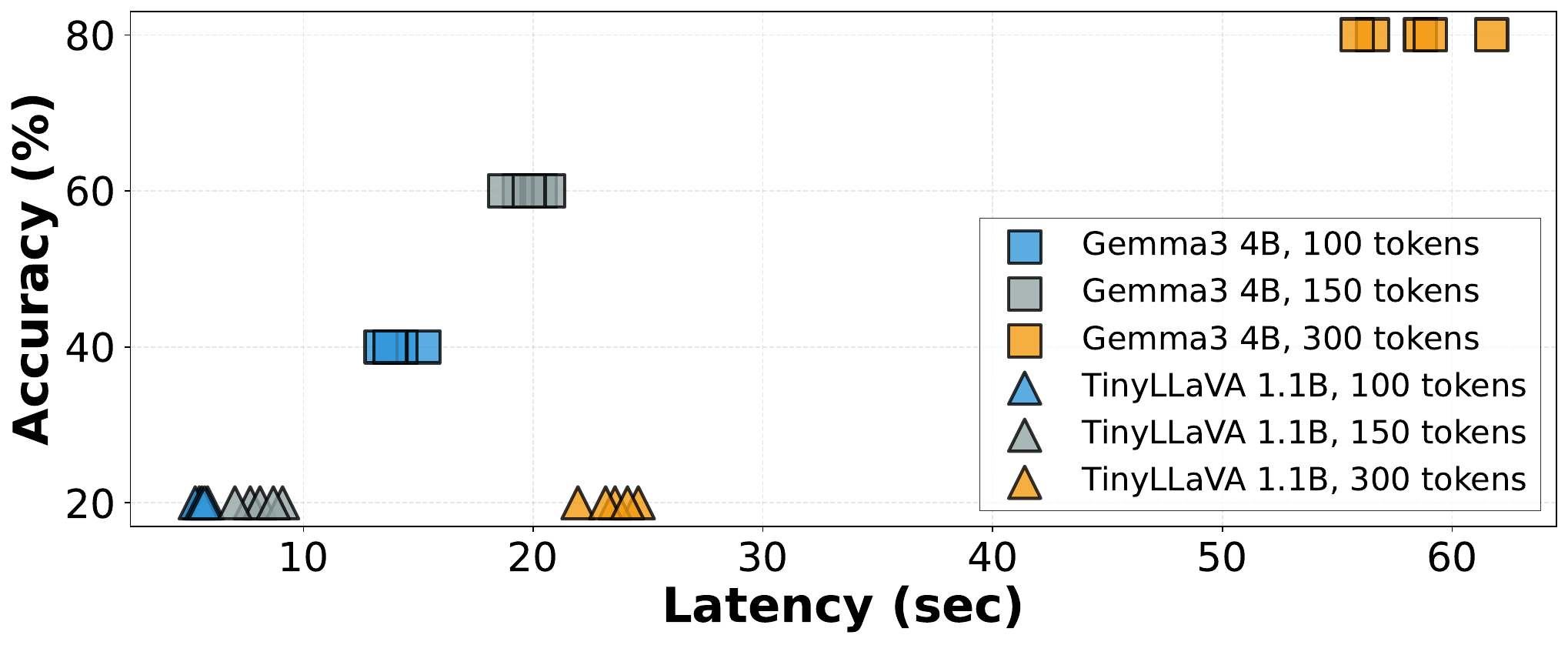}
\vspace{-2ex}
\caption{Trade-off between decision accuracy and latency for varying VLMs. Top left is better.}
\vspace{-2ex}
\label{fig:model_size_acc_latency}
\end{figure}

Figure~\ref{fig:model_size_acc_latency} presents the accuracy–latency tradeoff across different model sizes and maximum token limits. Due to the hardware constraints of the Jetson Orin NX, the largest deployable model is roughly 4B parameters; thus, we use Gemma3 4B as the primary VLM backbone~\cite{team2024gemma}. For comparison, we also evaluate TinyLLaVA (1.1B), one of the few sub-4B VLMs capable of processing visual input and a commonly used lightweight LLaVA variant.

The results imply that sufficient model capacity is crucial for reasoning performance. TinyLLaVA~\cite{zhou2024tinyllava} was unable to generate valid navigational decisions, even when provided with the same images and prompts as Gemma3 4B. Its accuracy plateaued around 20\% across all tested token limits (100, 150, 300), indicating that a 1.1B parameter budget is likely insufficient to support the necessary reasoning capabilities.

In contrast, Gemma3 4B achieved substantially higher performance and exhibited meaningful variation with token length, reaching its peak accuracy at a maximum token limit of 300 and with reasonable performances at 150. These findings confirm that a 4B-class model represents the minimum viable capacity for reliable reasoning in navigation settings.


\begin{figure*}[!t]
\centering
\includegraphics[width=0.95\linewidth]{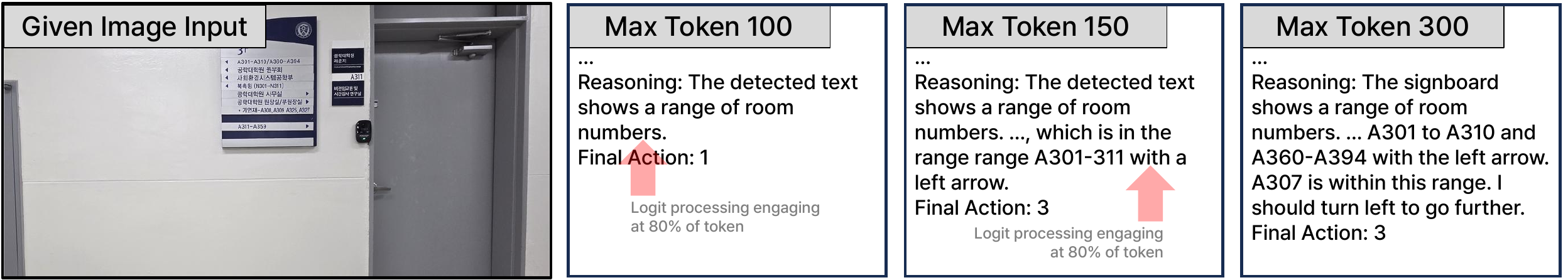}
\vspace{-2ex}
\caption{Comparison of reasoning outputs by max token length}
        \label{fig:model_reasoning_text_compare}
\end{figure*}

\subsubsection{Maximum Token Length}

Figure~\ref{fig:model_reasoning_text_compare} illustrates how the reasoning details produced through \textit{Additional Thinking} varies with different maximum token limits. Our qualitative evaluation reveals that the lightweight VLM ideally requires 300 tokens to fully articulate a descriptive phase necessary for grounding its reasoning. Notably, with this 300-token limit, the model typically reached a conclusion within 80\% of the available capacity in most scenarios, thereby avoiding any logit processing interventions.

However, this using 300 tokens imposes a significant computational burden, resulting in excessive latency, impractical for real-time deployments. As a result, we set a limit of 150 tokens, identifying this as the minimum operational threshold required to permit sufficient reasoning and derive accurate results while mitigating latency constraints.

\section{Discussion}
\label{sec:Discussion}

Based on our experiences in designing \system{}, we outline discussion points that can catalyze interesting future research.

\subsection{Advanced Scene Understanding}

Compact VLMs (e.g., 4B) often struggle with spatial and distance cues~\cite{chen2024spatialvlm, hong233d, jatavallabhula23conceptfusion} that are essential for contextual navigation. \system{} mitigates this by delegating simple cases to System One and supplying augmented spatial and textual descriptions to System Two, but this indirect prompt-fusion approach increases system complexity and does not reliably align segmentation or OCR outputs with the VLM latent space. Our experiments revealed cases where the VLM failed to interpret correct segmentation/OCR results when they were injected only as text. While Figure~\ref{fig:overall_performance}(a) shows that augmentation improves accuracy, robust operation ultimately requires integrating spatial and textual features directly into the visual encoder~\cite{medhi24targetprompting, zhang2024llavar}, a broader implication for all small, on-device VLMs deployed in real-world environments.

\subsection{Advanced Maneuvering Opportunities}

While \system{}'s Execution Module, based on VP tracking and PID control, provides reliable actuation in common indoor scenarios, its fluidity in dynamic and complex environments has not yet been validated. A lightweight reinforcement learning (RL) policy could serve as a local controller~\cite{tai2017virtual, omer24semantic}, taking \system{}’s high-level actions as targets while autonomously managing obstacle avoidance and smooth motion. Memory profiling on the Jetson Orin NX shows that \system{} occupies $\sim$13\,GB of the available 16\,GB, leaving sufficient headroom ($\sim$2\,GB) for inference-only RL policies (typically requiring <500\,MB). This indicates that a hybrid VLM–RL controller is feasible within the current hardware constraints.

A related limitation is the absence of an explicit obstacle-avoidance or safety module. The robot currently relies on VP tracking and coarse spatial segmentation, which is adequate in structured indoor spaces but insufficient for handling dynamic obstacles such as pedestrians or moved furniture. Since neither System One nor System Two is designed to guarantee collision-free behavior, unexpected environmental changes may result in unsafe motion.

For robust deployment in human-shared environments, \system{} will require additional safety features via depth sensing, collision buffers, or RL-based controls, to override or adjust actions in emergencies~\cite{baek2025ai}. Integrating such a module would maintain real-time responsiveness while satisfying essential safety requirements for embodied AI systems.




\subsection{Dual-Process Architecture Scalability}

The dual-process architecture, fast intuitive control paired with slower semantic reasoning, is broadly applicable beyond navigation. For manipulation tasks, System Two could handle high-level reasoning (e.g., identifying a target object or grasp pose), while System One performs repetitive or reactive motions through visual servoing or inverse kinematics~\cite{hutchinson1996tutorial}. Likewise, the decoupled nature of the architecture enables concurrent multi-tasking: System One can maintain locomotion while System Two processes a user request or analyzes visual information in parallel. This mirrors human behavior of intuitively navigating through an environment while reasoning about separate tasks, and highlights the scalability of \system{} toward general embodied intelligence~\cite{duan2022survey}.

\section{Conclusion}
\label{sec:Conclusion}

In this work we introduced \system{}, a hybrid navigation framework that bridges the gap between the semantic reasoning capabilities of Vision-Language Models (VLMs) and the strict real-time requirements of mobile robots. By instantiating the Dual Process Theory, \system{} decouples fast perceptual intuition (System One) from slow deliberative reasoning (System Two), enabling compact on-device VLMs to be invoked only when necessary. Real-world evaluations on a real robotic platform show that this design reduces total travel time by 60\% compared to VLM-based baselines and improves navigation success rates from 48.2\% to 64.3\% through Spatial \& Textual Information Augmentation. These results demonstrate that robust, low-latency VLM-based navigation is achievable on affordable hardware without sacrificing navigational decision quality. Future extensions will incorporate lightweight reinforcement learning schemes for smoother and safer execution, and generalize the dual-process design to multi-tasking and manipulation, advancing the development of practical, human-aware embodied AI systems.


\balance
\bibliographystyle{ACM-Reference-Format}
\bibliography{references, eis-lab}

\end{document}